\documentclass[12pt,letterpaper]{article}

\usepackage[margin=1in]{geometry}
\usepackage{times}
\usepackage{setspace}
\usepackage{tcolorbox}

\usepackage[utf8]{inputenc}
\usepackage[T1]{fontenc}
\usepackage{graphicx}
\usepackage{amsmath}
\usepackage{amssymb}
\usepackage{url}
\usepackage{hyperref}
\usepackage[authoryear]{natbib}
\usepackage{subcaption}

\usepackage{tabularx}

\usepackage{booktabs}
\usepackage{multirow}
\usepackage{array}

\usepackage{enumitem}
\usepackage{listings}
\usepackage{xcolor}
\usepackage{float}
\usepackage{orcidlink}

\floatstyle{plain}
\newfloat{listing}{tbp}{lol}
\floatname{listing}{Listing}

\graphicspath{{figures/}}

\hypersetup{
    colorlinks=true,
    linkcolor=black,
    citecolor=black,
    urlcolor=black
}

\title{A Dual-Helix Governance Approach Towards Reliable Agentic Artificial Intelligence for WebGIS Development}

\author{
    Boyuan Guan$^{1}$\orcidlink{0000-0002-4244-5011} \and 
    Wencong Cui$^{1,}$\thanks{Corresponding author: Wencong Cui (\texttt{wecui@fiu.edu}). Telephone: (305) 348-6443.} \orcidlink{0000-0001-6579-3724}\and 
    Levente Juhász$^{2}$\orcidlink{0000-0003-3393-4021}
}

\date{%
    \small $^{1}$Geographic Information Systems Center, Florida International University, Miami, FL, USA; \texttt{bguan@fiu.edu}, \texttt{wecui@fiu.edu}\\
    $^{2}$Geospatial Analytics, Technology and Open Research Lab, University of Florida, Fort Lauderdale, FL, USA; \texttt{levente.juhasz@ufl.edu}\\[2ex]
}

\begin{document}



\begin{titlepage}
\thispagestyle{empty}

\vspace*{0.35in}

\begin{center}
{\Large\bfseries
A Dual-Helix Governance Approach Towards Reliable Agentic Artificial Intelligence for WebGIS Development\par}
\end{center}

\vspace{0.25in}
{\normalsize
\textbf{Running title:} Dual-Helix Governance for WebGIS\par}

\vspace{0.30in}
\begin{center}
{\normalsize
\textbf{Boyuan Guan}$^{1}$\orcidlink{0000-0002-4244-5011} \\
\textbf{Wencong Cui}$^{1,*}$\orcidlink{0000-0001-6579-3724} \\
\textbf{Levente Juhász}$^{2}$\orcidlink{0000-0003-3393-4021}
\par}
\end{center}

\vspace{0.18in}
\begin{center}
{\small
$^{1}$ Geographic Information Systems Center, Florida International University, Miami, FL, USA \\
\texttt{bguan@fiu.edu}; \texttt{wecui@fiu.edu} \\
\vspace{0.08in}
$^{2}$ Geospatial Analytics, Technology and Open Research Lab, University of Florida, Fort Lauderdale, FL, USA \\
\texttt{levente.juhasz@ufl.edu}
\vspace{0.12in}
$^{*}$ \\ Corresponding author: Wencong Cui (\texttt{wecui@fiu.edu})
\par}
\end{center}

\vfill

\begin{tcolorbox}[
    colback=orange!8,
    colframe=orange!60!black,
    fonttitle=\bfseries,
    title={Preprint — Not Peer Reviewed},
    left=6pt, right=6pt, top=4pt, bottom=4pt
]
\small This is a preprint version of a manuscript under review in
\textit{Transactions in GIS}. The content may be revised following peer-review. Readers are advised to verify whether this is the most current version.
\end{tcolorbox}
\vspace{4pt}

\vspace{0.18in}

\vspace*{0.35in}
\end{titlepage}

\begin{abstract}
WebGIS development requires consistency, yet agentic AI often fails due to LLM context constraints, forgetting, stochasticity, instruction failure, and adaptation rigidity. We propose a dual-helix governance framework reframing these as structural problems rather than capacity deficits. Using a 3-track architecture (Knowledge, Behavior, Skills) and a persistent knowledge graph, it stabilizes execution by externalizing facts and enforcing protocols. Validation shows a governed agent successfully refactored a legacy WebGIS codebase (reducing cyclomatic complexity and improving maintainability), roughly halved trial-to-trial output variance relative to static prompting in a controlled experiment, and prevented common infodemic mapping errors in a 5-condition COVID-19 cartography ablation study. Operationalized via the open-source AgentLoom toolkit, this externalized governance provides the stability necessary for production-level geospatial engineering.
\end{abstract}


\newpage

\section{Introduction} \label{section-intro}

\subsection{The Reliability Challenge in WebGIS Development} \label{section-intro-challenge}

Consistency, reproducibility and domain-specific accuracy are some of the most important factors in all WebGIS development tasks. Practitioners or developers are required to navigate complex technology stacks, integrating specialized spatial libraries such as Mapbox GL JS, Leaflet or turf.js, while adhering to institutional and industry standards, accessibility requirements and a whole range of other considerations. This presents a challenge which is further compounded by a well-documented curriculum gap in GIS education. That is, GIS and geospatial professionals are not necessarily trained in modern computing sciences and software engineering principles. A review of 210 GIS and geography degree tracks in the U.S. found that only 10\% required a programming course for graduation \citep{bowlick_computer_2017}. Furthermore, existing GIS programming instruction often focuses narrowly on scripting within proprietary software environments rather than broader software engineering principles \citep{bowlick_course_2020}. This results in a workforce that may lack the cyber literacy necessary to create and navigate complex WebGIS development projects \citep{shook_cyber_2019}. Similarly, software engineers, developers and programmers traditionally trained in these disciplines are not necessarily skilled in understanding geographic principles, geospatial data handling or cartographic communication. This mismatch can result in faulty web maps rendering data on ''Null Island'' \citep{juhasz_i_2022} or simply in ineffective web maps unable to fulfill their promise \citep{mooney_mapping_2020}. Developing well-intended and functional WebGIS applications that adhere to both software engineering and geospatial best practices therefore requires a unique mix of two distinct skillsets. However, we note that this is not a GISCience-specific problem and the gap is known in general research software engineering as well \citep{cosden_research_2022, goth_foundational_2025}.

Most recently, artificial intelligence (AI) systems capable of planning, executing and iterating on complex tasks emerged, often called Agentic AI systems. They offer potential assistance for knowledge-intensive work, such as software development, even in the geospatial domain \citep{mai_opportunities_2024, hu_geoai_2019, li_giscience_2025}. However, when applied to WebGIS contexts, current agentic approaches show significant reliability gaps and therefore may not be the solution to the current needs of the GIS workforce. Just like how \citet{heaton_claims_2015} observe that many scientific software developers acquire their development knowledge from other scientific developers who also lack formal training, reliance on unvalidated agentic systems can lead to similar issues. In experimenting with agentic AI solutions for production WebGIS projects, we encountered systematic failures stemming from five core LLM limitations. Current models struggle with long-context comprehension degradation (C1) \citep{liu_lost_2024} and cross-session forgetting (C2) over extended development cycles. Additionally, inherent output stochasticity (C3) and frequent instruction-following failures (C4) preclude the enforcement of mandatory domain standards, such as strict cartographic rules or coordinate reference system consistency \citep{zhou_instruction-following_2023}. Finally, standard mitigation strategies like fine-tuning exhibit adaptation rigidity (C5), requiring resource-intensive, opaque updates rather than auditable, immediate behavioral corrections \citep{ouyang_training_2022}.


These gaps are symptoms of a fundamental structural mismatch: the absence of externalized governance mechanisms. Notably, a recent comprehensive survey of agentic reasoning identifies governance of autonomous agents as a cross-cutting open problem, arguing that ``agentic systems introduce new risks due to long-horizon planning, persistent memory, and real-world action execution'' and calling for ``governance frameworks that jointly address model-level alignment, agent-level policies, and ecosystem-level interactions under realistic deployment conditions'' \citep{wei_agentic_2026}. This confirms that the limitations we observe are not idiosyncratic to our setting but reflect recognized, field-wide challenges. Therefore, we frame the identified reliability challenges as a knowledge governance problem that cannot be resolved by model capability alone. We argue that reliability in production WebGIS development requires externalized governance structures that persist knowledge, enforce constraints, and stabilize execution. To address this, we propose a dual-helix governance approach that stabilizes agentic execution through two orthogonal axes: knowledge externalization and behavioral enforcement. In this paper, we describe how the Dual-Helix represents the structural stabilization of the agent, and how these two governance axes, combined with skills (validated, reproducible workflows), form a 3-track architecture that directly addresses limitations outlined above. This solution provides a methodological pathway for achieving what \citet{li_giscience_2025} identify as the five key functions of an autonomous GIS entity: decision-making, data preparation, data operation, memory-handling, and core-updating.

\subsection{Significance of this Study and Paper Organization} \label{section-intro-significance}

Rather than proposing a new learning algorithm or benchmark, this paper presents a production-grounded architectural design validated through two case studies: a real-world GIS deployment and a research-based web cartographic task. To evaluate the proposed framework, we define operational reliability as an agent's ability to complete complex, multi-step WebGIS development tasks with minimal human intervention during the execution phase, while preserving functional correctness and architectural consistency across sessions. This paper makes three primary contributions:
\begin{itemize} 
    \item \textbf{Conceptual Framework:} We propose a dual-helix governance approach for adapting agentic AI to WebGIS development, reframing reliability as a structural problem of externalized governance rather than model-internal capability.
    \item \textbf{Architectural Design:} We provide a methodological path for constructing knowledge graphs that encode WebGIS design patterns, behavioral constraints, and validated workflows as persistent, version-controllable artifacts.
    \item \textbf{Empirical Validation:} Through two independent case studies we provide empirical evidence to validate that governance structure is the primary driver of domain-specific reliability.
\end{itemize}

The remainder of this paper is structured as follows. Section~\ref{section-background-review} reviews related work in GeoAI and identifies the fundamental reliability gaps in existing agentic systems that necessitate a governance approach. Section~\ref{section-dual-helix} introduces the proposed dual-helix governance framework and details its operationalization through the 3-track architecture and dual-role separation. Section~\ref{section-futureshorelines} and Section~\ref{section-case-study-covid} present two case studies: the FutureShorelines refactoring case study, and a COVID-19 web mapping case study, respectively. The discussion of the results, their implications and the limitations of this study are described in Section~\ref{section-discussion}. Finally, Section~\ref{section-summary-futurework} summarizes the work and provides directions for future research.

\section{Background and Related Work} \label{section-background-review}

\subsection{The Evolution of Autonomous GIS and Informational AI Strategies} \label{section-background-evolution}

Geospatial AI (GeoAI) represents the integration of AI with geographic information science (GIScience). In addition to predictive modeling, generative and autonomous frameworks are gaining traction recently \citep{li_autonomous_2023, mai_opportunities_2024}. This evolution is characterized by the emerging paradigm of Autonomous GIS, which seeks to create systems capable of planning and executing complex spatial workflows with minimal human intervention \citep{li_giscience_2025}. Unlike traditional GIS software that acts as a passive tool for human-led analysis, autonomous systems leverage Large Language Models (LLMs) to bridge the gap between natural language intent and computational execution \citep{mai_opportunities_2024}.

Previous milestones demonstrated that even early versions of mainstream LLM chatbots could pass introductory GIS examinations \citep{mooney_towards_2023}, while newer generations successfully interpret spatial concepts and generate geospatial code \citep{hochmair_correctness_2024}. The field has gradually and quickly progressed towards functional "mapping assistants" \citep{juhasz_chatgpt_2023} and task-specific agents through frameworks such as MapAgent, that introduced hierarchical structures for geospatial reasoning \citep{hasan_mapagent_2024}. Another example, ShapefileGPT enables automated vector data manipulation \citep{lin_shapefilegpt_2024}. Recent efforts have further specialized these capabilities, with agents emerging for urban modeling \citep{li_urbangpt_2024}, remote sensing \citep{talemi_agentic_2026}, and automated cartographic design \citep{wang_cartoagent_2025}. \citet{weghe_opportunities_2025} further characterize this evolution and emphasize the transition from predictive models to generative and agentic frameworks. Beyond the frameworks mentioned before, \citet{ameen_review_2026} and \citet{zufle_comprehensive_2026} in two systematic reviews also identify a growing ecosystem of task-specific intelligent agents for working with spatio-temporal data and complex spatial reasoning. This confirms GIScience's growing interest in agentic workflows.

While these advances are promising, significant barriers remain. Geographic data structures, scale dependencies, and domain semantics create unique challenges for geospatial applications of AI \citep{xing_challenges_2023}. Even state-of-the-art models struggle with tasks requiring deep geographic context, such as complex topological reasoning \citep{ji_foundation_2025}. As the field continues to refine GeoAI, \citet{li_geoai_2024} emphasize that these systems must be grounded in predictability, interpretability, reproducibility, and social responsibility.

However, operationalizing these principles in agentic systems remains an open challenge. Current GeoAI agents are increasingly execution-capable, as demonstrated by their ability to generate executable geospatial workflows through concept transformations \citep{bao_spatial-agent_2026}, utilize geospatial libraries such as \textit{Geopandas} and \textit{Rasterio} \citep{akinboyewa_gis_2025}, and automate map generation \citep{li_autonomous_2023, juhasz_chatgpt_2023}. Nevertheless, we argue they remain fundamentally governance-deficient: a gap that extends well beyond GIScience. \citet{wei_agentic_2026} identify governance as one of six open problems in agentic AI, noting that existing benchmarks and guardrails ``mainly focus on short-horizon behaviors, leaving planning-time failures and multi-agent dynamics underexplored.'' We define governance-deficiency as a structural mismatch where a system can perform a task but cannot reliably adhere to the mandatory rules, standards, or long-term architectural consistency required for professional engineering. For WebGIS, this includes non-negotiable requirements like specific coordinate transformations for visualization, the use of approved libraries (e.g., \textit{Leaflet.js} vs \textit{OpenLayers}) and design choices (e.g. raster or vector tiles), and efficient data transmission protocols. 

This deficiency stems from a reliance on informational strategies to bridge the gap between model capability and domain requirements. Prompt engineering, the strategic design of input text to optimize output \citep{oxford_english_dictionary_prompt_2025}, serves as a primary yet manual interface for directing behavior. Other internal strategies include Chain-of-Thought (CoT) prompting, which enables models to decompose complex tasks into logical subunits \citep{wei_chain--thought_2022}, such as calculating topological relationships or multi-step routing. To reduce inherent stochasticity, researchers have also deployed Retrieval-Augmented Generation (RAG) architectures to ground generations in external facts \citep{lewis_retrieval-augmented_2020}, such as retrieving GDAL/OGR API documentation to ensure syntax accuracy. For example, prompt engineering can be used to constrain a model to output specific WebGIS configurations, such as forcing a \textit{MapLibre GL JS} implementation with a specific vector tile source. CoT prompting allows an agent to decompose a complex site-suitability analysis into discrete spatial operations, such as reprojecting layers, generating 500m buffers around transit hubs, and performing a geometric intersection with zoning polygons. To address hallucinations in code generation, RAG architectures can retrieve specific \textit{ArcPy} or \textit{PyQGIS} API documentation to ensure the agent uses correct method signatures for version-specific geospatial libraries.

Despite their utility, these strategies often fail in professional WebGIS engineering because they are advisory rather than systematically enforced. Prompt engineering is short-lived, residing only within a single prompt's context. Similarly, while RAG architectures excel at dynamically retrieving accurate facts (e.g. \textit{ArcPy} syntax or organizational standards), they remain fundamentally informational. They cannot structurally enforce compliance, such as mandating a specific coordinate reference system across multiple files. Instead, the retrieved text serves merely as context that the model may probabilistically ignore as task complexity grows. Our emphasis on a governance-based architecture directly addresses this limitation and aligns with the vision of Autonomous GIS, defined as a system capable of self-generating, self-organizing, self-verifying, self-executing, and self-growing \citep{li_giscience_2025, li_autonomous_2023}.

\subsection{Knowledge Graphs and the Reliability Ceiling}

Knowledge graphs (KGs) represent a structured approach to representing domain knowledge through nodes and edges, serving as a critical bridge between unstructured LLM outputs and formal geographic information. In GeoAI, KGs are utilized to ground model outputs to reduce hallucination \citep{pan_unifying_2024}, enable graph-enhanced retrieval to perform urban spatial reasoning \citep{chen_georag_2026}, and encode domain semantics for spatial question answering \citep{mai_se-kge_2020}. This research lineage draws directly from traditional GIScience work on Geo-Ontologies, which historically provided the semantic foundation for spatial data interoperability \citep{Agarwal01052005}. However, while traditional ontologies were often developed as top-down, static hierarchies, modern agentic workflows require a shift toward dynamic, bottom-up knowledge structures that evolve alongside complex project lifecycles.

Current approaches typically use these graph structures as static or task-specific resources for retrieval and orchestration rather than as active, persistent structures for behavioral control. While these frameworks successfully stabilize geospatial reasoning by grounding natural language in spatial information theory \citep{bao_spatial-agent_2026}, they remain essentially informational. They provide mechanisms for a single execution, but lack a mechanism to govern professional behavior or preserve project history across long-cycle projects such as WebGIS development. As a result, these approaches do not solve the fundamental structural mismatch that leads to a 'reliability ceiling' in production-level WebGIS. We argue that true operational reliability requires a transition from informational guidance to a persistent governance substrate that enforces non-negotiable standards, such as CRS reprojection, documentation consistency, and architectural patterns, as mandatory constraints rather than suggested steps. This lack of externalized governance makes current systems prone to the core limitations ($C1$--$C5$) identified in Section 1.1. Table \ref{tab:gaps} synthesizes these gaps, highlighting why current strategies fall short of the requirements for production-level geospatial engineering, essentially necessitating a move beyond execution-capability toward the Dual-Helix governance framework discussed in Section 3.

\begin{table}[h]
\centering
\small
\begin{tabular}{p{4.5cm}p{4.5cm}p{6.5cm}}
\toprule
\textbf{Approach} & \textbf{What It Addresses} & \textbf{Governance Gap} \\ \midrule
Prompting optimization & Single-task quality & Instructions are advisory; fails at $C2$ and $C4$ \citep{zhou_instruction-following_2023}. \\
RAG / Vector DB & Information retrieval & Lacks behavioral enforcement ($C4$) or validated workflows ($C3$). \\
Existing GIS agents & Narrow tasks (e.g., mapping) & Lacks long-term project memory ($C2$) or auditable adaptation ($C5$). \\
Fine-tuning (SFT/RLHF) & Model capability & Cycles are slow, opaque, and non-auditable ($C5$). \\ \bottomrule
\end{tabular}
\caption{Gaps in existing approaches relative to the five core LLM limitations identified in Section~\ref{section-intro-challenge}} 
\label{tab:gaps}
\end{table}

In addition, these informational approaches treat LLMs as static thinkers, but \citet{li_giscience_2025} argue that the most challenging yet critical goal for the field is self-growth, i.e. the ability to learn from successful and failed attempts to improve future performance. Our Dual-Helix approach directly targets this static capability by implementing a persistent substrate that serves as an evolving project memory, effectively bridging the gap between a model's training data and the real requirements of professional GIS engineering.

\section{Dual-Helix Governance Approach for WebGIS Development} \label{section-dual-helix}

\subsection{Conceptual Framework} \label{section-dual-helix-framework}

We propose a dual-helix governance approach to address the structural mismatch between standard LLM capabilities and the requirements of production-level WebGIS development. This approach reframes the challenge of operational reliability (see Section~\ref{section-intro-significance}) as a knowledge governance problem that cannot be sufficiently addressed by model capability alone. In this context, the "dual-helix" denotes the stabilization of a single agentic system through two orthogonal and co-evolving axes of control: Knowledge Externalization and Behavioral Enforcement. An agentic system governed by the dual-helix framework still relies on a LLM as the engine, however, the agent's knowledge, skills, and behaviors are stabilized by the framework.

\subsubsection{The Two Governance Axes}

\begin{figure}[htbp]
\centering
\includegraphics[width=\textwidth]{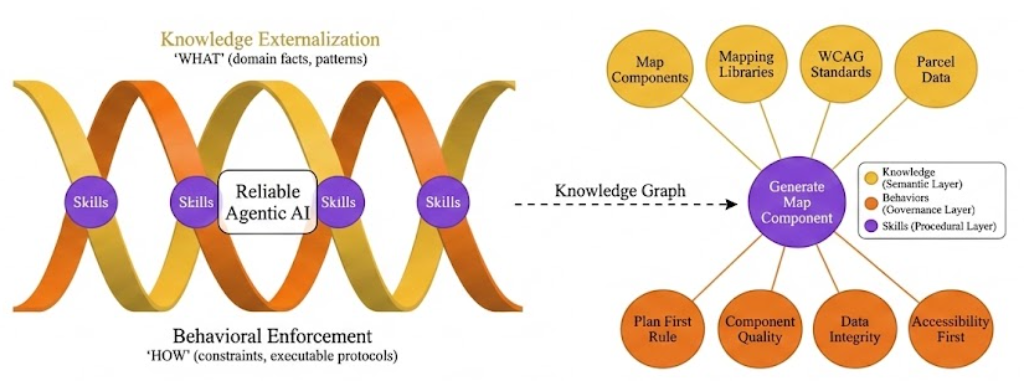}
\caption{The Dual-Helix Governance Framework for Reliable Agentic GeoAI. The framework stabilizes execution through two orthogonal axes: persistent Knowledge Externalization and enforceable Behavioral Enforcement connected via validated Skills.}
\label{fig:conceptual-framework}
\end{figure}

The metaphor of a "dual-helix" is intentionally chosen to emphasize that these two mechanisms (knowledge and behavior) are not merely parallel features, but are structurally intertwined and co-evolving. Just as strands of DNA are bound together by base pairs, the Knowledge axis (facts/patterns) and the Behavior axis (rules/constraints) are continuously interlocked through the self-learning cycle (Section~\ref{section-dual-helix-learning}); new project discoveries feed back into behavior updates, which in turn dictate what knowledge must be externalized next. As illustrated in Figure~\ref{fig:conceptual-framework}, these two axes act as a structural foundation that compensates for the inherent stochasticity and memory limitations of the underlying model.

\begin{itemize}
    \item \textbf{Axis 1: Knowledge Externalization:} This axis provides an auditable domain memory by shifting project-specific facts, architectural patterns and discovered contexts out of the ephemeral attention mechanism and into a persistent, version-controlled knowledge graph. By externalizing this information, the system directly addresses the constraints of long-context comprehension ($C1$) and cross-session forgetting ($C2$).
    \item \textbf{Axis 2: Behavioral Enforcement:} This axis introduces executable protocols that govern the agent's actions. Unlike traditional prompting, which is purely informational and advisory, behavioral enforcement makes compliance mandatory. By requiring the agent to validate its plans against these protocols before execution, we mitigate instruction-following failures ($C4$), such as the accidental violation of CRS standards that would render data to Null Island.
\end{itemize}

In a sense, Axis 1 provides the "What" (the foundational domain facts and patterns), while Axis 2 mandates the "How" (the executable constraints and protocols). Together, they form a co-evolving feedback loop where the agent's "execution capability" is structurally stabilized by a mandatory governance structure. This dual-helix interaction ensures that as the system performs complex tasks, it remains anchored in a persistent institutional memory that survives session boundaries and architectural complexity.

\subsection{The 3-track Architecture} \label{section-dual-helix-architecture}

\begin{figure}[htbp]
\centering
\includegraphics[width=\textwidth]{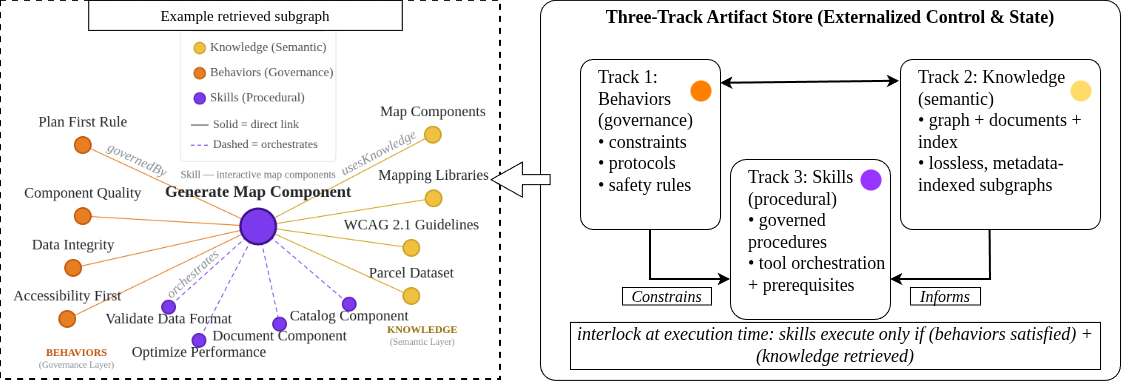}
\caption{The 3-track architecture operationalizing the dual-helix approach. Track 1 (Knowledge) and Track 2 (Behaviors) represent the governance axes, while Track 3 (Skills) provides stabilized execution patterns.}
\label{fig:3-track}
\end{figure}

To operationalize the conceptual framework described above, we translate the Knowledge and Behavior axes into a functional, 3-track architecture that serves as the implementation substrate directly targeting the identified constraints ($C1$--$C5$). The three tracks are illustrated in Figure~\ref{fig:3-track}. The architecture is implemented as a unified Knowledge Graph (KG) where every governance artifact (i.e., a rule, fact, or workflow) is stored as a version-controlled node. These nodes follow a specific taxonomy to manage different granularities of information (Table~\ref{tab:knowledge-types}). 

Programmatically, the KG is structured as a collection of JSON documents, which allows the autonomous agent to traverse the hierarchical relationships. Each node definition includes a semantic description used for contextual retrieval and a relative path linking to a local Markdown file containing the expanded, human-readable instructions. An illustrative example of a conceptual knowledge node from the COVID-19 mapping case study (see Section~\ref{section-case-study-covid}) is shown in Listing~\ref{lst:node-example}.

\begin{listing}
\caption{Example JSON definition of a knowledge node mitigating cartographic misinformation. \textit{Note:} Extended examples of Knowledge, Skill, and Behavior nodes, along with their corresponding Markdown files, are provided in Supplementary Material~\ref{supplementary-kg-definitions}}
\label{lst:node-example}
\begin{lstlisting}
{
   "id": "webgis:bias-ethics:infodemic",
   "type": "document",
   "path": "docs/webgis-developer/bias-ethics/visual-misinformation.md",
   "title": "Preventing Visual Misinformation in Pandemic Mapping",
   "parent": "webgis:bias-ethics",
   "description": "Applying Mooney & Juhasz (2020) principles to counter alarming visualizations."
}
\end{lstlisting}
\end{listing}

This structural approach aligns with the community goals for Autonomous GIS, which envision systems that move beyond passive toolsets to become active "artificial geospatial analysts". We specifically map our framework to the five autonomous behaviors defined by \citet{li_giscience_2025}: the Skills track enables self-generating and self-executing workflows via validated templates, while the Behaviors track enforces self-verifying structural validation. Additionally, the hierarchical Knowledge Graph and role separation facilitate self-organizing resource management, and the Knowledge track drives self-growing adaptation through a persistent feedback loop

\begin{table}[h]
\centering
\small
\begin{tabular}{lll}
\toprule
\textbf{Type} & \textbf{Purpose} & \textbf{Example} \\ \midrule
Category & Organize related concepts & \texttt{gis-research:ai-gis-integration} \\
Concept & Abstract domain ideas & \texttt{knowledge:event-driven-architecture} \\
Document & Specific artifacts with content & \texttt{docs/patterns/modular-design.md} \\ \bottomrule
\end{tabular}
\caption{Knowledge node types in the 3-track architecture}
\label{tab:knowledge-types}
\end{table}

\textbf{Track 1 (Knowledge)} operationalizes the Knowledge Externalization axis, serving as the agent's institutional memory. By grounding the agent in verifiable domain facts rather than potentially incorrect training data, this track enables the storage of technology stacks, design patterns, and project-specific contexts. This externalization addresses $C1$ (long-context limitations) by offloading structural data from the model's ephemeral attention mechanism and addresses $C2$ (cross-session forgetting) through a standardized initialization protocol that restores project context from the graph. Furthermore, the track enables auditable self-learning: i.e., as the agent discovers new patterns, it persists them as new nodes, providing immediate adaptation without retraining ($C5$).

\textbf{Track 2 (Behaviors)} serves as the executable implementation of the Behavioral Enforcement axis, functioning as the governance layer that systematically constrains agent actions through executable protocols rather than mere suggestions. Unlike implicit rules embedded in advisory prompts, behaviors are documented, versioned nodes in the KG that determine task completion. Each behavior node contains metadata, including a priority level (Critical, High, or Medium) to determine enforcement strictness, and links to the skills it governs. Before executing any skill, the agent must retrieve all governing behaviors and validate that intended actions comply with these requirements, ensuring that critical constraints like WCAG accessibility or CRS integrity cannot be accidentally ignored.

\textbf{Track 3 (Skills)} represents the intersection where knowledge and behavior axes meet to form stabilized workflows. Each skill defines its required inputs, expected outputs, and the protocols it must satisfy to create reproducible execution patterns. When a skill is invoked, the agent interlocks the relevant knowledge nodes and behavior constraints to execute the process with full context. This structured execution mitigates the inherent stochasticity of the underlying model ($C3$), ensuring that same inputs combined with the same governance structure yield consistent, professional-grade architectural outcomes.

\subsection{Role Separation as a Stabilization Mechanism} \label{section-dual-helix-stabilization}

\begin{figure}[htbp]
\centering
\includegraphics[width=0.9\textwidth]{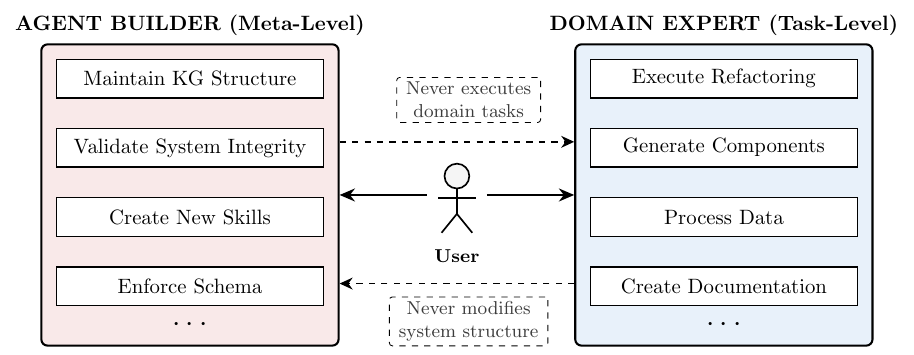}
\caption{Role separation as implementation mechanism. The Agent Builder maintains system structure while the Domain Expert executes project tasks. Explicit role switches prevent context contamination.}
\label{fig:dualrole}
\end{figure}

To ensure the integrity of the governance axes during long-horizon tasks, we employ role separation as another protective implementation pattern. This mechanism is not a defining theoretical component of the dual-helix itself, but rather a structural safeguard designed to prevent context contamination, a phenomenon where LLMs conflate different responsibilities or lose track of governance protocols during extended interactions. As illustrated in Figure~\ref{fig:dualrole}, the system can toggle between two distinct operational states:

\begin{itemize}
    \item \textbf{Agent Builder (Meta-level):} This role is responsible for maintaining the KG structure, validating system integrity, creating new skills, and enforcing schema compliance. The Agent Builder operates strictly during the "Build Phase."  The Agent Builder can itself be implemented as an agent powered by an LLM (see COVID-19 case study in Section~\ref{section-case-study-covid}), but the role can also be fulfilled by a human, as was the case in the first case study by one of the authors to provide quality control (Section~\ref{section-futureshorelines}). The human fulfilling this role acts as a systems architect defining the legal boundaries and governance protocols of the project, not as a pair-programmer. The Builder never executes domain-specific tasks or writes project code, ensuring that the governance of the system remains focused purely on architectural health.
    
    \item \textbf{Domain Expert (Task-level):} This role operates at the project level during the "Execution Phase" to perform the actual labor: refactoring, generating code, processing geospatial data, and creating documentation. The Expert operates autonomously within the boundaries set by the Builder. While the Expert utilizes a "plan-first" approach that allows for human-in-the-loop oversight (e.g. a human approving a proposed plan), this oversight is limited to safety gatekeeping. The human can author zero lines of project code and make zero technical correction if desired (see Case Study 2 in Section~\ref{section-case-study-covid}). Furthermore, the Expert role is prohibited from modifying the system governance structure, acting as a specialized worker that follows the protocols established in the Knowledge and Behavior tracks.
\end{itemize}

By enforcing explicit role switches, the framework externalizes concerns that LLMs otherwise conflate, preserving the reliability of the dual-helix axes across long development cycles.

\subsection{Self-Learning Mechanism} \label{section-dual-helix-learning}

Unlike static retrieval systems, the dual-helix approach enables the agent to evolve through a structured self-learning mechanism. In this context, learning refers to the acquisition, documentation, and persistence of knowledge discovered during project work, rather than updates to model parameters. As illustrated in Figure~\ref{fig:selflearning}, this mechanism follows a verifiable five-step cycle:

\begin{figure}[htbp]
\centering
\includegraphics[width=0.95\textwidth]{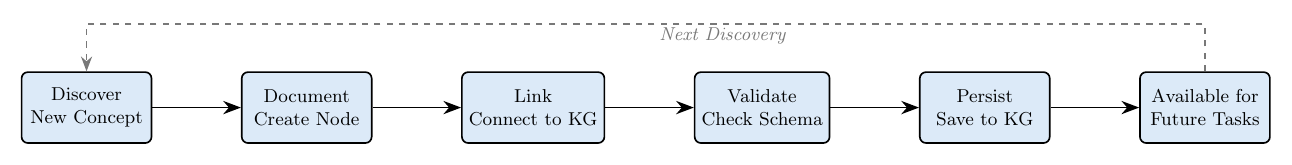}
\caption{The conceptual self-learning mechanism. As the agent performs tasks, it discovers, structures, links, validates, and persists new project context as auditable graph nodes. Figure~\ref{fig:selflearning-experiment} illustrates how this cycle is operationalized in the experimental workflow.}
\label{fig:selflearning}
\end{figure}

\begin{enumerate}
    \item \textbf{Discovery:} The agent identifies new concepts, library-specific patterns, or institutional requirements during execution (e.g., discovering a specific geographic dataset and its schema).
    \item \textbf{Structuring:} The discovery is formalized as a structured knowledge node with typed metadata (e.g., node category, method signatures, and parent references).
    \item \textbf{Linking:} The new node is programmatically connected to the existing graph hierarchy in the KG (Category or Concept).
    \item \textbf{Validation:} The system programmatically checks that the new node meets established JSON schema requirements, subject to human (Agent Builder) review.
    \item \textbf{Persistence:} The updated graph is saved, making the knowledge immediately available for future sessions without re-explanation.
\end{enumerate}

This mechanism directly addresses $C5$ (Adaptation Rigidity) by allowing the system to adapt to project-specific requirements in real-time. Because these changes are persistent graph artifacts, the learning is auditable, version-controlled, and immediately reversible if necessary. The self-learning aspect of the framework is further illustrated through the first case study described in Section~\ref{section-futureshorelines-refactor-outcome}.

\subsection{Open-Source Implementation} \label{section-dual-helix-reliability}

The framework described in this section is implemented as the open-source system \textit{AgentLoom} available at \url{https://doi.org/10.5281/zenodo.17561541} \citep{guan_agentic-ai_2025}. The system utilizes the dual-role architecture consisting of a standardized Agent Builder for system maintenance and a custom Domain Role for specific task execution (Section~\ref{section-dual-helix-stabilization}). The core governance substrate is implemented through hierarchical KGs following a JSON-based schema. These graphs are designed for full connectivity, i.e., every node, with the exception of the root, is required to define a parent field to ensure a single-rooted tree structure. Semantic relationships across the Knowledge, Behavior, and Skills tracks are managed via a \texttt{links} object, which defines how behaviors govern specific skills or enforce domain facts. To ensure operational reliability across complex workflows, the framework uses "specification-based generation", i.e., it creates fresh components from technical requirements rather than using templates. It also maintains persistent state through a "phase memory" system that explicitly saves context variables between development stages.

While this paper focuses on the theoretical governance approach in WebGIS development and its validation through two distinct case studies, the full software architecture and deployment workflow of AgentLoom will be discussed in detail in a forthcoming implementation study. In the following section, we demonstrate how this governed architecture enables the autonomous refactoring of a complex legacy WebGIS application.
\section{Case Study 1: FutureShorelines Project} \label{section-futureshorelines}

\subsection{Project Background} \label{section-futureshorelines-background}

\begin{figure}[htbp]
\centering
\includegraphics[width=0.85\textwidth]{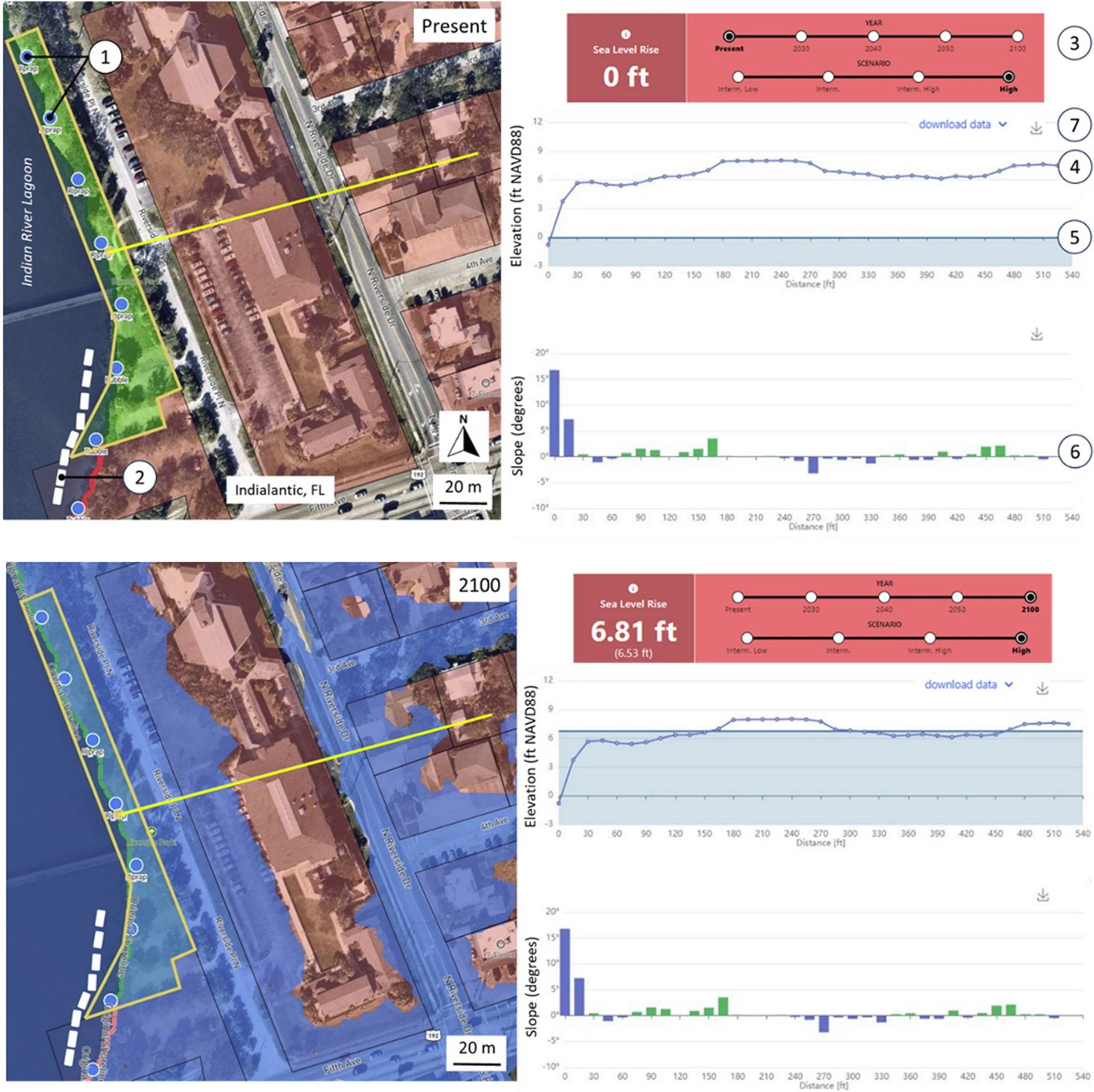}
\caption{The FutureShorelines decision support tool showing a typical use-case for planning living shoreline installations \citep{parkinson_future_2024}}
\label{fig:futureshorelines}
\end{figure}

The FutureShorelines project is a web-based geospatial decision support system (DSS) designed to facilitate climate-resilient coastal management. Originally developed for the Indian River Lagoon (IRL) in Florida, the tool integrates high-resolution LiDAR elevation models with NOAA sea-level rise (SLR) scenarios to characterize future conditions of submergence and shoreline translation \citep{parkinson_future_2024}. The platform employs a reproducible inundation mapping methodology based on free and open-source software for geospatial (FOSS4G) workflows \citep{juhasz_beyond_2023}. Through a co-production framework, the platform provides practitioners with interactive visualizations of future inundation extents, shoreline topography, and property ownership to identify optimal locations for nature-based stabilization structures, such as living shorelines. Figure~\ref{fig:futureshorelines} illustrates the application for planning living shoreline installations. For effective decision making, this task requires reliable information on property ownership (green = public, red = private), the extent of inundated areas in the future, shoreline topography, etc. More information about the project and the web application can be found at \url{https://futureshorelines.fiu.edu/}.

While the initial deployment focused on living shoreline site selection, the FutureShorelines framework has since been applied to other coastal management questions. For example, \citet{parkinson_sea_2025} utilized the tool to simulate the vulnerability of estuarine ecosystem services, including the impact of submergence on motorized boat ramps, spoil islands, and seagrass distribution. However, adapting the initial deployment to new geographic regions or research requirements presents a significant technical barrier for scientific teams. The current case study focuses on the transition from the original IRL implementation to a new deployment at the Rookery Bay National Estuarine Research Reserve (RBNERR). This new project's overarching goal is to provide a vulnerability assessment tool for the reserve, which goes beyond the scope of the original application developed for the IRL that only considered site-suitability for nature-based solutions. 

The primary challenge lies in the software's architectural history. The legacy system was developed as a monolithic 2,265-line JavaScript application. While scientifically robust, the codebase lacks the modularity required for rapid adaptation to the RBNERR use case. Modifying the tool to support different data schemas, local tidal datums, and visualization requirements traditionally requires extensive manual refactoring, which is often resource-prohibitive for small research teams. We utilize this transition from IRL to RBNERR as a testbed for our agentic AI workflow, evaluating its ability to govern the refactoring of a monolithic scientific application into a modular, production-grade GIS framework, which is the first step towards deploying the FutureShorelines application to the RBNERR use-case.

\subsection{The Technical Debt of the Existing System}

The original FutureShorelines WebGIS tool (\url{https://futureshorelines.fiu.edu/app/}) prioritized the rapid dissemination of coastal vulnerability data over long-term software architectural standards. It was created by one developer, and one of the project principal investigators who had no formal training in software development, both of them working on the project part-time. This setup is common in funding-constrained academic settings. While effective for its initial deployment in the Indian River Lagoon (IRL), the codebase of this application is characterized by significant technical debt by modern software engineering standards (See Section~\ref{section-intro-challenge} about the GIS curriculum gap) that complicates scalability and adaptation for different use-cases and geographic areas. The legacy system consists of a 2,265-line monolithic JavaScript file (comprising 1,086 logical source lines of code (SLOC)), global variables, no formal automated testing, hardcoded configuration values and minimal documentation. This monolithic architecture presents three primary challenges:

\begin{itemize}
    \item \textbf{Tight integration of specialized libraries:} Spatial analysis routines (via turf.js), map rendering (via Mapbox GL JS), and interactive data visualization (via ECharts) are interconnected within the same functional blocks. For example, a logic calculating distance from the shoreline is deeply nested within user interface (UI) event listeners. This makes it difficult to update the SLR scenario logic for the RBNERR deployment without risking failures in the map or chart managers.
    \item \textbf{Global state vulnerability:} The application relies on global variables to manage map state, temporal sliders and tidal datums. In the context of LLM-assisted development, this lack of encapsulation can trigger long-context limitation ($C1$). Standard LLMs often fail to track variable mutations across such a large, unstructured file, leading to the introduction of ghost variables or broken state logic during refactoring attempts.
    \item \textbf{Hardcoded domain logic:} Parameters such as coordinate bounds, local tidal datums, and specific SLR offsets are hardcoded directly into the application logic. Abstracting these into a configuration layer is a prerequisite for moving the tool from the IRL to RBNERR, yet doing so manually is resource-prohibitive.
\end{itemize}

For a typical research team or scientist-developer, manual refactoring of this codebase into a modern system was estimated to require 80–120 hours. This refactoring task exhibits all five LLM limitations from Section~\ref{section-intro-challenge}. That is, the codebase exceeds comfortable context lengths ($C1$), development spans multiple sessions over days ($C2$), consistent refactoring patterns are required ($C3$), organizational standards must be enforced ($C4$), and project-specific knowledge must be learned and retained ($C5$).

\subsection{Code Modernization via AgentLoom: Dual-Helix Framework at Work} \label{section-futureshorelines-refactor}

We applied the dual-helix approach to transform the monolithic FutureShorelines codebase into a modular, maintainable architecture. The goal was to decouple the core coastal vulnerability modeling logic from the UI, so that the system could be adapted to new geographic regions, updated SLR scenarios, and alternative use-cases without breaking existing functionality. To achieve this, the 3-track structure was initialized with domain-specific knowledge and behavioral constraints.

We used the Cursor agentic Integrated Development Environment (IDE) to implement AgentLoom with \texttt{gpt-5.2} as the underlying execution model. Following the dual-role architecture (Section~\ref{section-dual-helix-stabilization}), a human researcher served as the Agent Builder, overseeing the governance structure and reviewing architectural plans, while the LLM operated as the Domain Expert executing project tasks.

The Knowledge track was initialized with domain knowledge nodes, for example with the detailed methodology of modeling coastal inundation described in \citep{juhasz_beyond_2023}, project context (i.e. project proposal) and other relevant information. Technical knowledge nodes provided the agent with patterns for the project's core dependencies (Mapbox GL JS, ECharts, turf.js), while architectural knowledge nodes described the target modular state. The Behavior track was configured with accessibility compliance requirements (WCAG 2.1 Level AA) and code quality constraints (500-line file limits, modular architecture requirements).

A central philosophy of this implementation was the "plan-first" rule, which operationalized the human-in-the-loop requirement. While this might appear in tension with the concept of fully "autonomous" GIS, we argue that autonomy in professional software engineering must prioritize reliable execution over the complete removal of oversight. In modern Continuous Integration/Continuous Deployment (CI/CD) environments, no developer (human or AI) pushes a 2,000+ line architectural refactoring to production without a senior architect reviewing the Pull Request.

Before the agent executed any implementation actions, it was mandated to generate a structured Refactoring Plan. This checkpoint acted as a structural safety gate and a method for human intervention during this case study. For example, the human overseeing the execution phase rejected the agent’s initial proposed plan to refactor the files in-place (i.e. override the original codebase), forcing the agent to autonomously rethink the architecture and output the modules to a completely new, sandboxed directory to protect the legacy code. The agent documented this modification and made it auditable through a revised Refactoring Plan (shown in Supplementary Material~\ref{supplementary-refactor}).

During the actual generation of the six ES6 modules, the human intervened 7 times. While the human authored zero lines of code, they acted as an active Quality Assurance (QA) tester (e.g. by pasting console errors back to the agent and catching a missing parcel layer) and pointing out UI regressions. Because the human actively guided the debugging process in this first case study to ensure the legacy system did not break, the observed reliability gains cannot be fully decoupled from human oversight. To isolate the framework's standalone reliability and address this exact limitation, a second, entirely autonomous experiment was conducted (Section~\ref{section-case-study-covid}).

\begin{figure}[htbp]
\centering
\includegraphics[width=\textwidth]{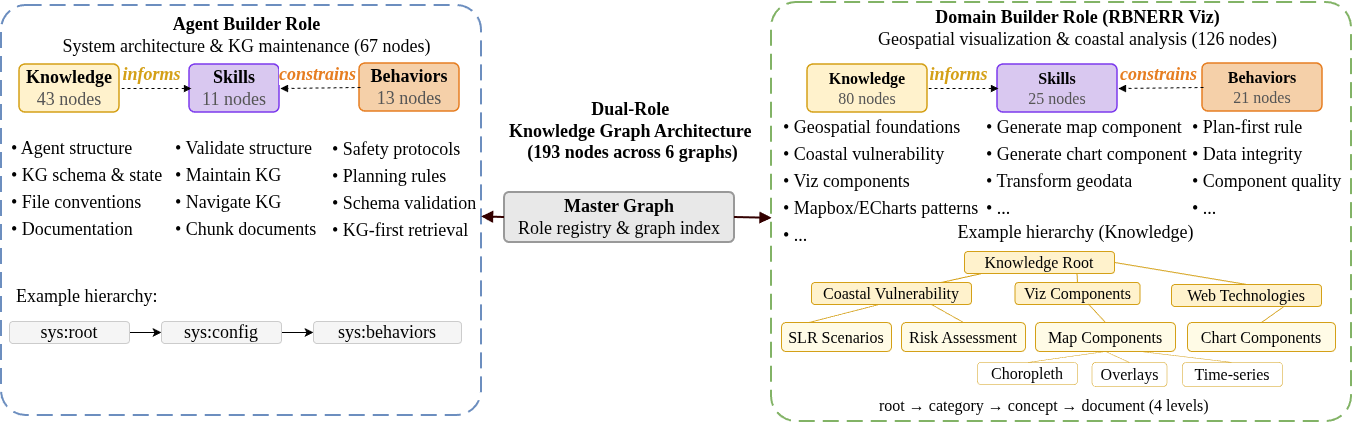}
\caption{Knowledge graph structure for the FutureShorelines project. The hierarchical organization includes domain knowledge (geospatial concepts, coastal vulnerability) and system knowledge (validation rules, file conventions).} 
\label{fig:kg}
\end{figure}

Figure \ref{fig:kg} shows the resulting KG substrate, illustrating how domain knowledge (geospatial concepts, coastal vulnerability) and system knowledge (validation rules, file conventions) are organized hierarchically.

Once the governance substrate was initialized and the research team approved the refactoring plan, the modernization proceeded in four phases over four development sessions spanning three days. In Phase 1, the agent parsed the 598-line \texttt{PROJECT\_BACKGROUND.md} document, extracting scientific methodology, sea-level rise scenarios, and institutional requirements into persistent knowledge nodes. This grounded the system in verifiable project facts rather than model-internal training data. In Phase 2, the agent analyzed the 2,265-line legacy monolithic code against the established behavioral requirements. During this phase, it identified unique project patterns such as custom vector tile fallback logic, which were formalized as persistent skill nodes via the self-learning mechanism and added to the Knowledge Graph. In Phase 3, following the "plan-first" rule, the agent refactored the monolithic codebase into six cohesive JavaScript ES6 modules (\texttt{config.js}, \texttt{mapManager.js}, \texttt{chartManager.js}, \texttt{dataManager.js}, \texttt{uiManager.js}, and \texttt{main.js}). The governance substrate rejected non-compliant plans (e.g., modules exceeding the 500-line limit) before execution. In the final phase (Phase 4), the agent generated technical documentation and validated the output against WCAG 2.1 Level AA accessibility standards, confirming that the modernized application met the professional engineering criteria established in the Behavior track.

\subsection{Outcomes of the Refactoring Task} \label{section-futureshorelines-refactor-outcome}

The refactoring transformed the 2,265-line monolith into six cohesive ES6 modules, eliminating global scope dependencies and establishing a unified configuration and event-driven communication pattern. The modular architecture also makes the codebase easier to maintain and adapt to new scenarios. For example, new charts quantifying seagrass distribution \citep{parkinson_sea_2025} can be implemented directly in the \texttt{chartManager.js} module.  

To quantify the structural improvements, we evaluated four standard software metrics: Logical Source Lines of Code (SLOC) \citep{jay_cyclomatic_2009}, Cyclomatic Complexity \citep{mccabe_complexity_1976}, the Maintainability Index \citep{oman_metrics_1992, microsoft_code_metrics_2024}, and JSHint Warnings. As Table~\ref{tab:code-metrics} demonstrates, the refactoring produced measurable improvements across all dimensions. The 49\% reduction in logical SLOC was achieved by eliminating redundant logic and modularizing shared utilities. The 51\% reduction in cyclomatic complexity reflects the decomposition of deeply nested legacy conditional logic into manageable, single-responsibility functions. Only one JSHint warning remained, indicating strong adherence to modern ECMAScript standards, and the 7-point increase in the maintainability index (on a 100-point scale) confirms the improved codebase.

\begin{table}[h]
\centering
\small
\begin{tabular}{lrrr}
\toprule
\textbf{Metric} & \textbf{Legacy State} & \textbf{Modernized State} & \textbf{Change} \\ \midrule
Logical SLOC          & 1,086  & 555   & $-49\%$ \\
Cyclomatic Complexity & 126    & 62    & $-51\%$ \\
Maintainability Index & 59  & 66 & $+7$ pts \\
JSHint Warnings       & 51     & 1     & $-98\%$ \\ \bottomrule
\end{tabular}
\caption{Code quality metrics before and after governance-guided refactoring.}
\label{tab:code-metrics}
\end{table}

We also measured the self-learning capabilities of the architecture through the growth of the project-specific knowledge graph. During the refactoring, the agent identified and formalized undocumented project contexts that were not part of the initial knowledge base. These were added as persistent graph nodes, and included for example specific vector tile fallback logic and delayed chart initialization for hidden containers. Table~\ref{tab:kg-growth} documents the growth in KG nodes throughout the refactoring task. The graph grew from 28 seed nodes to 126 nodes, showing that the agent externalized and persisted discovered contexts without human technical intervention. A total of 98 new nodes were created autonomously. To ensure quality, these autonomously generated nodes were subject to human review (via the Agent Builder role), which confirmed that they accurately captured project patterns without introducing redundancy or hallucinations. This represents a transition from general model knowledge to a specialized memory. These new nodes are auditable, version-controlled and are immediately available for subsequent development sessions.

\begin{table}[h]
\centering
\small
\begin{tabular}{lrrr}
\toprule
\textbf{Graph Component} & \textbf{Initial Nodes} & \textbf{Final Nodes} & \textbf{Growth} \\ \midrule
Project Knowledge & 15 & 80 & 433\%  \\
Project Skills & 8 & 25 & 213\%  \\
Project Behaviors & 5 & 21 & 320\%  \\ \midrule
\textbf{Total Substrate} & \textbf{28} & \textbf{126} & \textbf{350\% } \\ \bottomrule
\end{tabular}
\caption{Growth of the KG substrate (governance structure) during the FutureShorelines refactoring project.}
\label{tab:kg-growth}
\end{table}

\subsection{Testing Operational Reliability: A Controlled Experiment} \label{section-futureshorelines-experiment}

To isolate the impact of the dual-helix governance structure from general model capability, we conducted a controlled experiment. The objective was to determine whether externalized governance improves the operational reliability of the agent, or if providing the same information in a traditional prompt is sufficient. This experiment was designed to test the framework as a whole.

Unlike the human-guided refactoring described above (Section~\ref{section-futureshorelines-refactor}), which involved active human oversight, this experiment evaluated the agents on a complex, 5-step WebGIS dashboard refactoring workflow designed to perform an equivalent transformation but operating in a fully autonomous mode without human-in-the-loop oversight. This workflow required the agents to independently navigate a 2,432-line monolithic legacy file to execute a sequence of five sequential steps. The workflow steps and prompts are described in detail as Supplementary Material~\ref{supplementary-workflow} and summarized below:

\begin{enumerate}
    \item Extracting information to create a centralized configuration module (i.e. \texttt{config.js})
    \item Migrating chart logic into a \texttt{ChartManager} class
    \item Refactoring map initialization into a \texttt{MapManager} class
    \item Refactoring UI interactions with accessibility support and
    \item Generating technical documentation
\end{enumerate}

\begin{figure}[htbp]
\centering
\includegraphics[width=\textwidth]{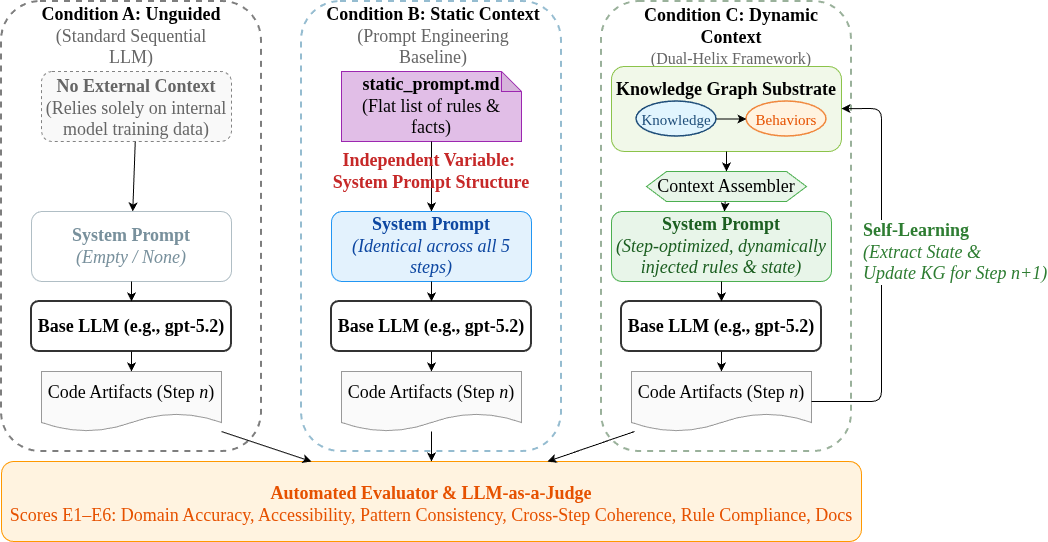}
\caption{Controlled experiment design isolating the system prompt structure. The conversation history, user prompt, and base LLM were identical across all conditions to ensure a fair comparison.}
\label{fig:experiment-design}
\end{figure}

To ensure reproducibility and eliminate human bias, the experiment was executed programmatically using a custom Python-based evaluation pipeline that interfaced directly with the LLM APIs. This automated framework orchestrated the multi-step conversations, maintained state, and executed evaluation checks without human intervention. To ensure a fair comparison and isolate the mechanism of system prompt structure, we tested three distinct conditions using the same underlying LLM (gpt-5.2). All conditions received the same user prompts (i.e. step instructions described in Supplementary Material~\ref{supplementary-workflow}) and conversation history, as well as the legacy codebase. The experimental design is illustrated in Figure~\ref{fig:experiment-design} and the tested conditions are described below:

\begin{itemize}
    \item \textbf{Condition A (Unguided Sequential):} The baseline agent operated with no external context, relying solely on its internal training, conversational history, the legacy codebase as well as user prompts describing step instructions.
    \item \textbf{Condition B (Static Context):} The agent was provided a comprehensive static system prompt containing all project background documents, domain facts, and accessibility rules. This represents the ceiling of traditional manual prompt engineering, ensuring the control group had access to the exact same information as the governed agent. Importantly, no token limit was imposed on any condition; all operated within the model's full context window. The key difference is therefore not token volume but prompt structure: the governed condition retrieved step-specific constraints dynamically, whereas the static condition provided all information in a single monolithic prompt. The per-step static system prompt (~4,000 tokens) was considerably larger than each dynamically assembled governance prompt (~1,400 tokens), confirming that the governed condition's advantage stems from the structural organization of constraints rather than a larger volume of injected context. The system prompt is given as Supplementary Material~\ref{supplementary-systemprompt-b}.
    \item \textbf{Condition C (Dynamic Context / Dual-Helix):} The agent operated within the full dual-helix framework. The system prompt was dynamically assembled at each step by querying the Knowledge Graph for step-specific constraints and injecting accumulated states from previous steps. An example dynamic prompt for Step 4 is given in Supplementary Material~\ref{supplementary-systemprompt-c}.
\end{itemize}

The experiment measured performance across six dimensions: Domain Accuracy ($E1$), Accessibility Compliance ($E2$), Pattern Consistency ($E3$), Cross-Step Coherence ($E4$), Rule Compliance ($E5$), and Documentation Accuracy ($E6$). Each dimension was scored on a 3-point scale (0 = non-compliant, 1 = partially compliant, 2 = fully compliant). The scoring utilized a hybrid approach in which deterministic checks served as the primary evaluation mechanism: requirements such as exact coordinate systems, layer IDs, and prohibited APIs were validated via automated regex and keyword checks. Qualitative dimensions such as cross-step coherence were assessed as a secondary measure using an LLM-as-a-judge framework utilizing the \texttt{gpt-5.2} model. Each condition was run across five independent trials; given the computational cost and latency of multi-step, multi-turn LLM execution (i.e. each trial comprising a full 5-step autonomous workflow), this sample size represents a practical compromise that nonetheless permits meaningful variance analysis. We calculate a cumulative score that represents the combined weighted performance, normalized to a maximum of 10.0. Criteria most relevant to operational reliability ($E4$ and $E5$) are assigned a weight of 1.5 while others contribute with a weight of 1.0. To quantify the stability of each approach, we calculated the standard deviation ($\sigma$) across all trials for each condition. In the context of production-grade software development, this metric serves as a proxy for operational reliability and a reduced variance indicates that the system produces predictable, stable results rather than stochastic, one-off successes. More detail about the evaluation is given in Supplementary Material~\ref{supplementary-eval}.

\begin{figure}[htbp]
\centering
\includegraphics[width=0.7\textwidth]{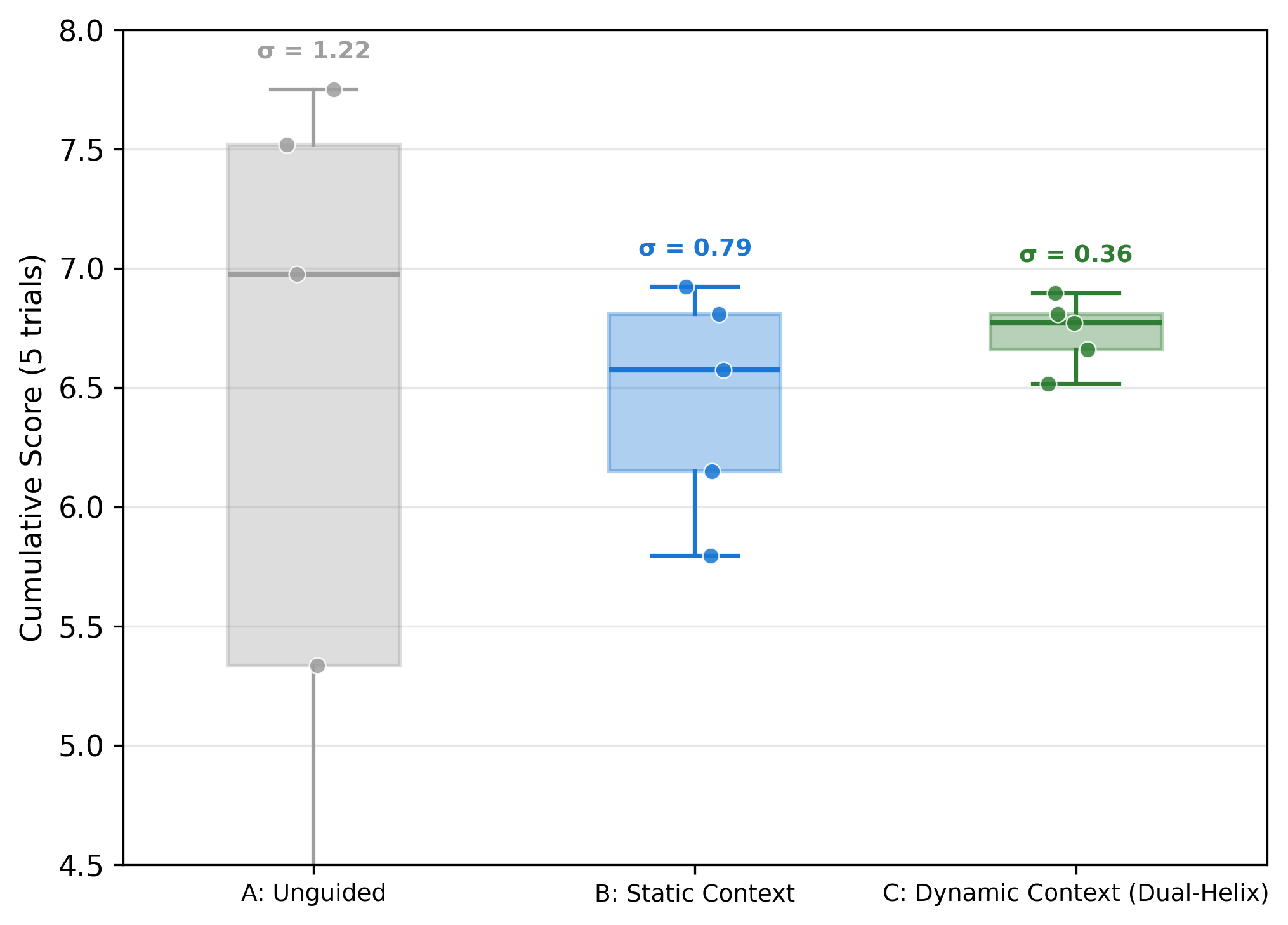}
\caption{Boxplots of trial-level consistency across conditions (gpt-5.2). The dual-helix approach (Condition C) substantially reduces variance}
\label{fig:reliability-variance}
\end{figure}

Unsurprisingly, both static and dynamic context engineering outperform the unguided approach. Results further suggest that governance structure increases operational reliability. Figure~\ref{fig:reliability-variance} shows boxplots of the cumulative score based on five trials for all three conditions. The standard deviation ($\sigma$) for each condition is also provided in the plot. While Condition B (Static Context) achieved a similar mean score (6.45) to Condition C (Dual-Helix)'s 6.73, it exhibited high trial-to-trial variance. In fact, Condition C (Dual-Helix) reduced the standard deviation by more than half ($\sigma = 0.36$ vs. $\sigma = 0.79$). Even though a Welch’s t-test showed that the difference in mean performance between the static (B) and dual-helix (C) conditions was not statistically significant ($t(5.18) = 1.60, p = 0.169$) due to the constrained sample size, the difference represents a small-to-medium effect size (Cohen's $d = 0.46$). More importantly, the primary value of the framework in this specific software engineering task is not necessarily raising the maximum capability ceiling, as the baseline LLM can occasionally achieve high scores probabilistically, but rather acting as an operational stabilizer. By reducing performance variance, the Dual-Helix architecture effectively transforms unpredictable, stochastic LLM outputs into more reliable, consistent engineering workflows.

This variance reduction has a concrete operational meaning. In practical software engineering, a high-variance LLM ($\sigma=0.79$) represents an operational liability because it dictates the worst-case performance of the agent. To quantify what a reduction of 0.43 in standard deviation means for the workflow, we look at the system's operational lower bounds. Under the static baseline (Condition B: $\mu=6.45$, $\sigma=0.79$), the expected lower bound of the agent's performance routinely drops into the critical failure zone (scores dropping toward $\approx 4.8$). In a real-world workflow, a developer cannot blindly merge code from a process with a lower bound this severe; the output is likely to contain catastrophic logic failures or architectural hallucinations that require the human to execute a full, manual rewrite (a high-cost intervention). By compressing the variance by 0.43 ($\sigma=0.36$) and raising the mean ($\mu=6.73$), the Dual-Helix framework raises the operational floor of the agent: the worst-case performance is tightly clustered around structural soundness (scores $\approx 6.0$). For a human developer, this translates directly to avoided interventions, as the worst-case scenario is elevated from a complete module rewrite to a standard code review (e.g., correcting minor syntax tweaks or variable names).

Furthermore, on metrics evaluating strict Rule Compliance ($E5$) (e.g., adhering to precise DOM manipulation limits and specific coordinate system requirements), Condition C (Dual-Helix) outperformed Condition B (Static Context), achieving a mean score of $1.66$ compared to $1.30$ for the baseline ($+27.7\%$ improvement). This pattern suggests that when domain constraints are embedded within a massive static prompt, they are frequently ignored as context grows; however, when constraints are enforced as dynamic, step-optimized behavioral protocols, compliance becomes more consistently maintained. Combined with the variance reduction findings, these results suggest that the dual-helix approach transitions agentic output from a probabilistic outcome to a more reliable engineering process.

\begin{figure}[H]
\centering
\includegraphics[width=0.95\textwidth]{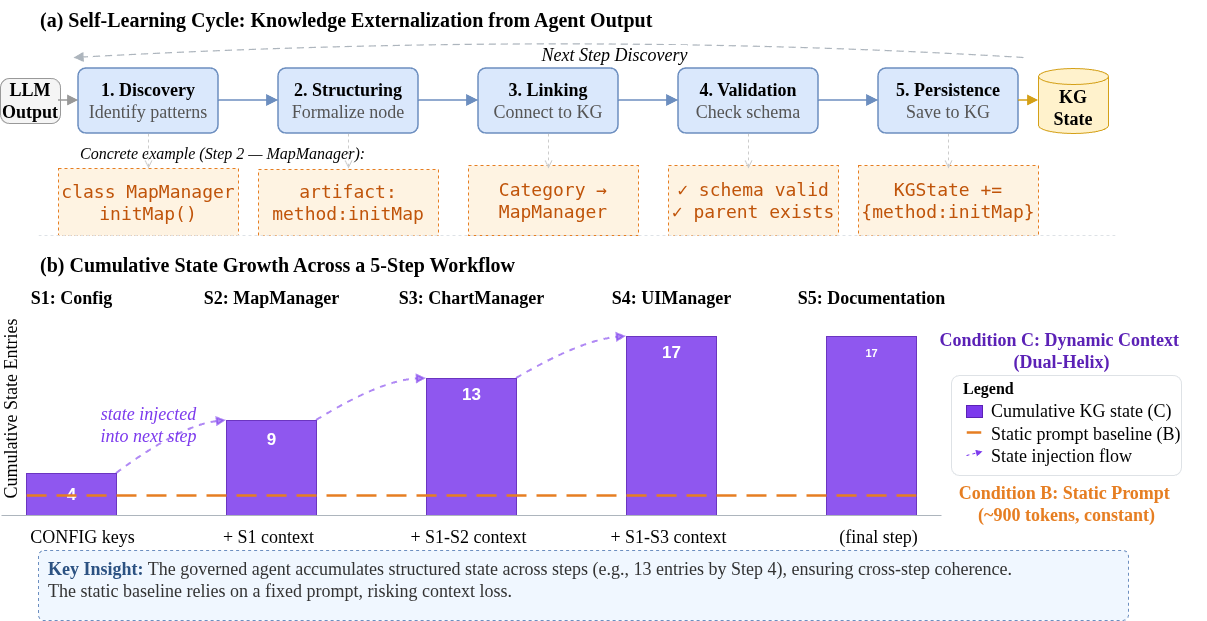}
\caption{(a) The 5-step Self-Learning Cycle for knowledge externalization, and (b) the resulting cumulative state growth across the five-steps of the controlled experiment for Condition C.}
\label{fig:selflearning-experiment}
\end{figure}

The operational reliability observed in the governed condition is supported by the autonomous self-learning cycle illustrated in Figure~\ref{fig:selflearning-experiment}. This figure operationalizes the conceptual five-step cycle introduced in Section~\ref{section-dual-helix-learning} (Figure~\ref{fig:selflearning}) within the experimental workflow. As shown in Figure~\ref{fig:selflearning-experiment}a, the agent first identifies architectural constituents during task execution (Discovery), such as recognizing that the \texttt{initMap()} method should belong to a \textit{MapManager} class. These discoveries are then formalized into typed state entries with structured metadata (Structuring), connected to existing graph categories (Linking), validated against the JSON schema (Validation), and saved to the KG state (Persistence).

Figure~\ref{fig:selflearning-experiment}b visualizes the resulting Cumulative State Growth across the experimental workflow. While the permanent KG contains immutable domain facts like SLR scenarios (see Table~\ref{tab:kg-growth}), this session-specific "KG State" serves as a dynamic memory that accumulates these discoveries. The "State injection flow" explicitly demonstrates how these accumulated discoveries (growing from 4 to 17 entries) are carried forward and injected back into the system prompt of each subsequent step. This recursive mechanism ensures the agent remains anchored in its own recent architectural decisions, effectively bridging the context gap between isolated model calls and mitigating the risks of cross-session forgetting and output stochasticity.

\section{Case Study 2: COVID-19 Web Mapping} \label{section-case-study-covid}

While the FutureShorelines case study described in the previous Section demonstrated the efficacy of the integrated architecture in a complex legacy environment, it did not isolate the independent contributions of the governance mechanisms. To address this, we designed the following five-condition ablation study that also demonstrates the framework's domain generalizability beyond legacy code refactoring by expanding the evaluation to automated cartographic design. 

\subsection{Cartographic Reliability during the COVID-19 Pandemic}

Specifically, we tasked an agent with generating a web-based geospatial visualization of COVID-19 data. The global SARS-CoV-2 pandemic was accompanied by a massive "infodemic," defined as an overabundance of information, both accurate and inaccurate, which severely complicated public access to trustworthy information \citep{tangcharoensathien_framework_2020}. The flexibility and accessibility of modern web-based mapping architectures allowed digital maps to become active vectors of this infodemic. Because modern web mapping requires little to no formal cartographic training, poor design choices often inadvertently create misinformation and opportunities for misinterpretation. In the context of LLM-generated maps, unguided agents may also fall into specific, predictable cartographic traps. \citet{mooney_mapping_2020} identify several prevalent errors in COVID-19 web mapping, such as a lack of data normalization, emotional color palettes, and inconsistent spatial aggregation. We use this research to evaluate the framework's capacity to structurally enforce these cartographic standards.

Furthermore, to demonstrate that externalized governance is a universal principle rather than an artifact of proprietary APIs and software utilized in the first case study, this case study was implemented in another agentic IDE, Google Antigravity. The generation task was executed utilizing \texttt{gemini-3.5-flash}. This setup points towards the Dual-Helix governance framework's software agnosticism.

\subsection{Ablation Experimental Design}

To isolate the specific mechanisms driving cartographic reliability, we designed a five-condition ablation study. The objective was to autonomously generate a complete, interactive COVID-19 WebGIS visualization component from  provided raw GeoJSON datasets consisting of point locations simulating confirmed COVID-19 cases, and polygon population data (census blocks).

Prior to task execution, the framework underwent a deliberate "Build Phase" orchestrated by the meta-level Agent Builder role. The agent was provided with the \citet{mooney_mapping_2020} manuscript detailing COVID-19 cartographic pitfalls and an activation prompt (detailed in Supplementary Material Listing~\ref{lst:activation-prompt}). From this unstructured text, the Agent Builder autonomously extracted and formalized the domain rules into a JSON graph substrate. This process generated the explicit semantic artifacts (e.g., \texttt{webgis-developer-knowledge-graph.json} and \texttt{webgis-developer-behaviors-graph.json}), populating knowledge and behavior nodes (see e.g. Listing~\ref{lst:node-example} in Section~\ref{section-dual-helix-architecture}). This demonstrates that the agent's governance structure is not a static, hardcoded artifact, but a dynamically generated, auditable institutional memory. The following conditions were implemented. These condition labels (A--E) are specific to this ablation study and differ from those used in the controlled experiment in Section~\ref{section-futureshorelines-experiment}; here, Condition~B denotes the fully governed framework, while Conditions~C, D, and E denote single-track ablations.

\begin{itemize}
    \item \textbf{Condition A (Unguided Baseline):} The base model operating with only the raw user prompt, relying entirely on internal training weights and the default IDE settings.
    \item \textbf{Condition B (Full Dual-Helix):} The framework operating with access to domain-specific cartographic facts built by the agent using \citet{mooney_mapping_2020}
    \item \textbf{Condition C (Knowledge Ablation):} The Full Dual-Helix framework operating without access to externalized domain knowledge
    \item \textbf{Condition D (Behaviors Ablation):} The Full Dual-Helix framework operating without access to externalized behavioral constraints
    \item \textbf{Condition E (Skills Ablation):} The Full Dual-Helix framework operating without access to externalized skills 
\end{itemize}

Each condition was initialized with a prompt describing the task (Supplementary Material~Listing~\ref{lst:task-prompt}). For conditions C D and E, the JSON files representing the corresponding KG as well as markdown documents referenced in these graphs were physically removed from the folder, along with any internal memory, context, and cached files in the IDE's environment to prevent access to these externalized artifacts for the ablation conditions. In addition, the agents in Conditions C, D and E were also instructed to not rebuild the ablated artifacts and fix the missing links in the knowledge graph. Each condition was then run three times. 

\subsection{Evaluation}

To quantitatively assess the cartographic reliability of the generated web maps, we developed a deterministic scoring rubric based on the primary failure modes identified by \citet{mooney_mapping_2020}. Each condition was evaluated across three independent trials using a compliance score (0 = Fail, 1 = Partially OK, 2 = Pass) capturing failures across the dimensions shown in Appendix Table~\ref{tab:rubric}. Across three independent runs, the evaluation scores from three independent evaluators (the authors) were averaged. The 15 solutions were randomized and labeled numerically so that evaluators did not know which condition produced the map they evaluated.

When evaluating applications with user-selectable parameters, scoring was based on the default state and the presence of contextual guidance. If an application offered weak options (e.g. raw counts, equal interval), but defaulted to sound methods and explicitly warned the user about the limitations of alternatives, it was not penalized.

\subsection{Results of the COVID-19 Mapping Case Study} \label{section-covid-results}

Inter-Rater Reliability was measured across the three independent evaluators (the authors). Given the ordinal nature of the scoring system and the small sample size ($n=3$ trials per condition), standard variance-based metrics heavily penalize minor subjectivity. Instead, we calculated exact and adjacent percentage agreement (scores within 1 point of each other). Across all 75 graded items, the evaluators reached an exact consensus 32\% of the time.  However, they achieved 91\% adjacent agreement (scores within 1 point of each other). Furthermore, to assess whether evaluators agreed on the relative quality of the generated maps even when their absolute baseline scores differed, we calculated the pairwise Spearman's rank correlation. The average Spearman's rank correlation across all evaluator pairs was $\rho = 0.60$, indicating a strong agreement on the relative ranking of the architectures.

In 9\% of cases (7 instances) where evaluators experienced disagreement (one evaluator scoring 0 while another scoring 2), the discrepancies were entirely confined to the Baseline and partially ablated conditions due to ambiguous or broken code outputs. Importantly, the maps generated by the full dual-helix condition achieved 100\% adjacent agreement, which suggests that the fully governed architecture produces outputs with consistent quality, eliminating evaluator discrepancy. 

\begin{figure}[H]
\centering
\includegraphics[width=0.65\textwidth]{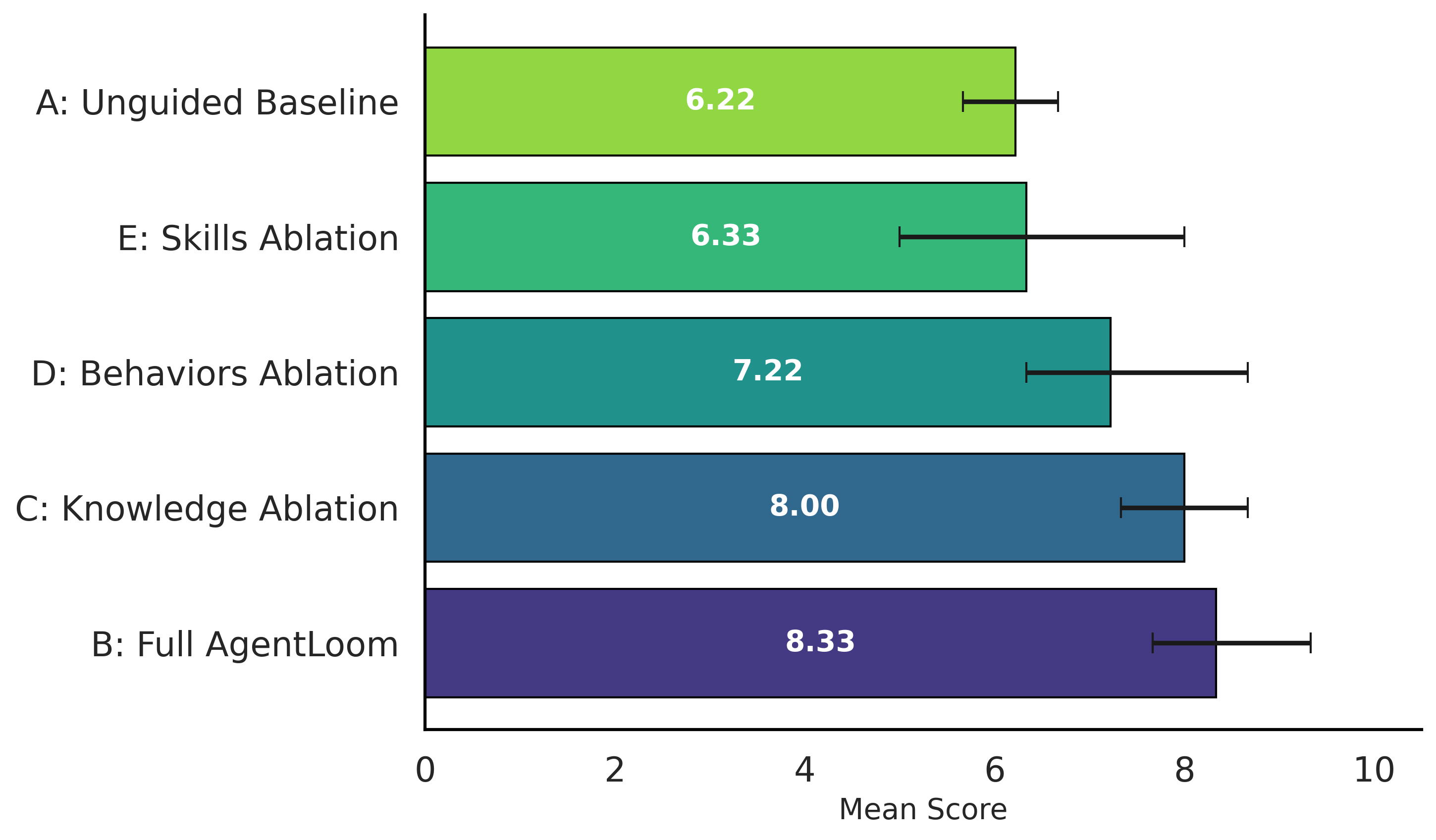}
\caption{Performance across 3 trials. Error bars represent minimum and maximum across trials for a condition}
\label{fig:covid-19-performance}
\end{figure}

The ablation study confirmed the efficacy of the Dual-Helix governance framework in preventing common COVID-19 web mapping pitfalls. When directly comparing the unguided Baseline to the Full Dual-Helix implementation, evaluators rated the governed AgentLoom architecture as superior to the Baseline 78\% of the time, and equal or superior 100\% of the time. Figure~\ref{fig:covid-19-performance} shows the average score for the five conditions across three independent trials. When evaluating the total cartographic reliability scores (max 10), Condition B (Full Dual-Helix) emerged as the best architecture, achieving an average total score of $8.33$. Condition A (Unguided Baseline) recorded the lowest performance across the study, scoring $6.22$. 

Removing any single track reduced performance, but the magnitude of the drop differed sharply across tracks, allowing us to rank their relative contribution for this cartographic task. Removing the Skills track (Condition E, $6.33$) caused the largest degradation, returning performance to essentially the unguided baseline level ($6.22$); removing the Behaviors track (Condition D, $7.22$) caused an intermediate drop; and removing the Knowledge track (Condition C, $8.00$) had the smallest effect, remaining within rounding of the Full Dual-Helix score ($8.33$). This orders the per-track contribution as Skills $>$ Behaviors $>$ Knowledge for this task, which we interpret as indicative rather than definitive given the small sample ($n=3$ runs per condition). Table~\ref{tab:descriptive_stats} shows the average scores across evaluation criteria for all conditions. A Kruskal-Wallis H-test across all five conditions yielded $H=6.675$ ($p=0.15$). While the differences did not meet the 95\% level confidence threshold for statistical significance primarily due to the limited sample size ($n=3$ runs per condition) and inter-evaluator variance, the large effect size ($\epsilon^2 = 0.48$) suggests a practical difference between the unguided and governed architectures. 

\begin{figure}[htbp]
    \centering
\includegraphics[width=0.95\textwidth]{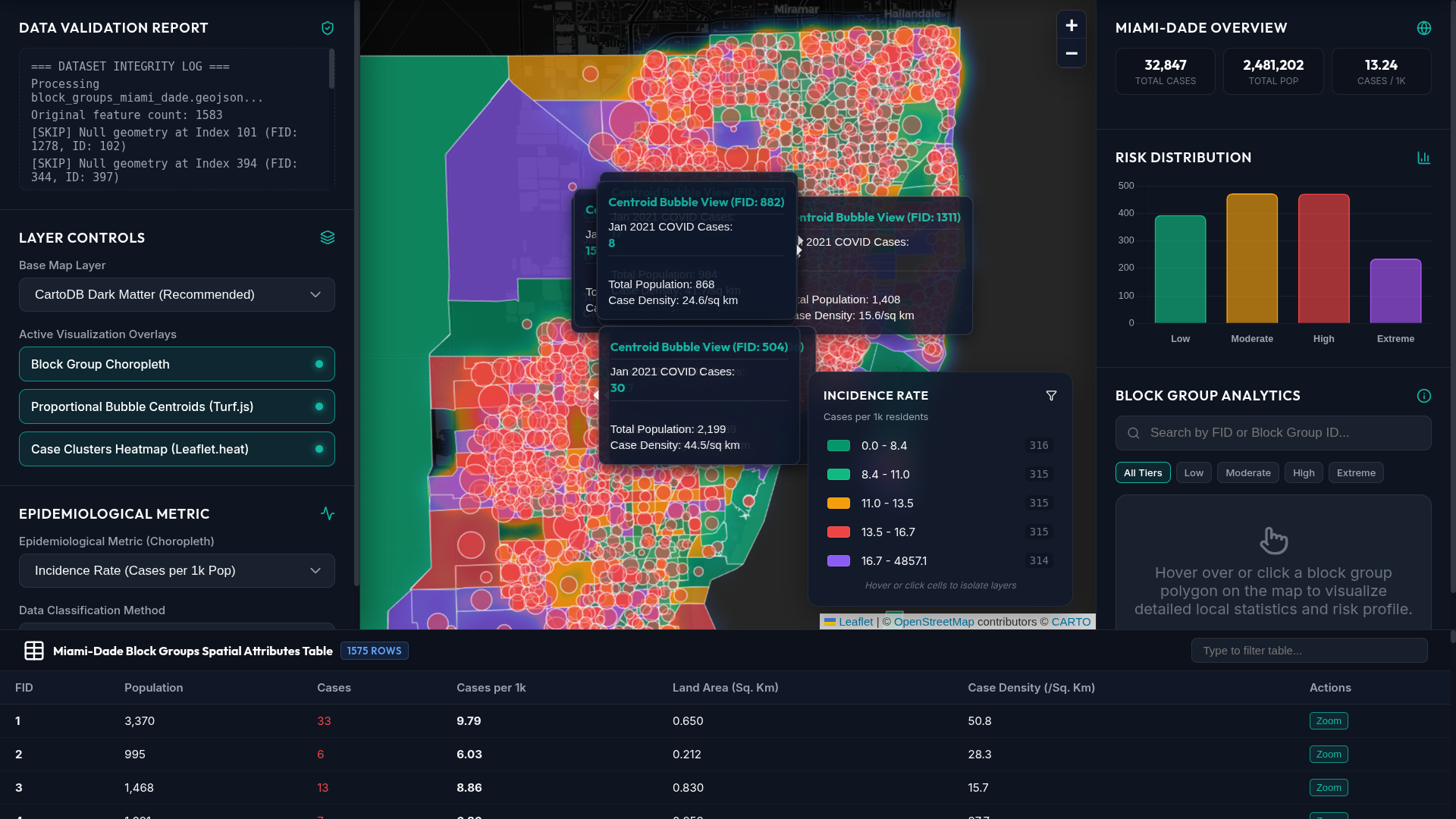}
    \caption{Output from Condition A (Baseline). The unguided LLM presents raw case counts without warning, and omits critical provenance elements such as data source attribution, and presents a visually cluttered interface with flawed functionality (e.g. overlapping pop-ups)}
    \label{fig:map_baseline}
\end{figure}

Breaking down the performance across the dimensions constructed based on \citet{mooney_mapping_2020} reveals how the agentic governance mitigated "infodemic" risks (Table~\ref{tab:descriptive_stats}):

\begin{itemize}
    \item \textbf{D1 (Aggregation \& Metrics):} While both the Baseline (1.67) and Full Dual-Helix (1.89) performed relatively well, the unguided model missed methodological warnings about offering raw counts, therefore occasionally generating misleading visualizations that simply mirrored population density rather than actual disease severity. In contrast, the governed agent reliably calculated and mapped normalized rates, mitigating this classic infodemic pitfall. We note that on this dimension the Knowledge- and Behaviors-ablation conditions ($2.00$ each) marginally exceeded the Full Dual-Helix score ($1.89$); we attribute this to the relative simplicity of rate normalization on this dataset, which the base model often handles even without the externalized Knowledge track, so that the Knowledge track's contribution instead surfaces on harder dimensions such as D4 (uncertainty and provenance documentation). 
    \item \textbf{D2 (Classification \& Symbology):} The fully governed agent outperformed (1.67) the baseline (1.44) in successfully choosing appropriate classification methods and perceptually uniform color palettes, whereas the baseline more frequently utilize flawed methods (see e.g. Figure~\ref{fig:map_baseline}).
    \item \textbf{D3 (Layer \& View Management) \& D5 (Technical Functionality):} The governance framework effectively structured the visual interface and coding process. Condition B (Full Dual-Helix) consistently outscored the Baseline in both preventing visual clutter through effective layer management (1.78 vs. 1.44) and successfully compiling interactive UI elements without console errors (1.67 vs. 1.33). 
    \item \textbf{D4 ((Spatial) Uncertainty):} The largest failure of the Baseline agent occurred in documenting spatial limitations and dataset metadata, scoring 0.33 out of 2.0. Unguided agents routinely failed to acknowledge methodological limitations or attribute data sources, which was identified a critical vulnerability in pandemic mapping. The full implementation (Condition B) improved this metric to 1.33 through explicitly mandated semantic behaviors (e.g., \texttt{behavior:audit-trail}).
\end{itemize}

\begin{table}[htbp]
\centering
\caption{Descriptive statistics of cartographic reliability across ablation conditions ($n=3$ runs per condition, averaged across 3 evaluators). Maximum total score = 10.0; Maximum dimension score = 2.0.}
\label{tab:descriptive_stats}
\renewcommand{\arraystretch}{1.2}
\begin{tabularx}{\textwidth}{@{} >{\hsize=1.2\hsize\raggedright\arraybackslash}X c c c c c c @{}}
\toprule
\textbf{Condition} & \textbf{Total Score} [min, max] & \textbf{D1} & \textbf{D2} & \textbf{D3} & \textbf{D4} & \textbf{D5} \\
\midrule
\textbf{A: Baseline}             & 6.22 [3, 8] & 1.67 & 1.44 & 1.44 & 0.33 & 1.33 \\
\textbf{B: Full Dual-Helix (AgentLoom)}      & \textbf{8.33} [7, 10] & 1.89 & 1.67 & 1.78 & \textbf{1.33} & 1.67 \\
\textbf{C: Knowledge Ablation}   & 8.00 [6, 9] & 2.00 & 1.56 & 1.78 & 1.11 & 1.56 \\
\textbf{D: Behaviors Ablation}   & 7.22 [6, 10] & 2.00 & 1.56 & 1.33 & 1.00 & 1.33 \\
\textbf{E: Skills Ablation}      & 6.33 [4, 9] & 1.89 & 1.22 & 1.11 & 0.89 & 1.22 \\
\bottomrule
\end{tabularx}
\end{table}

\begin{figure}[htbp]
    \centering
    
    \begin{subfigure}[b]{0.7\textwidth} 
        \centering
        \includegraphics[height=7cm]{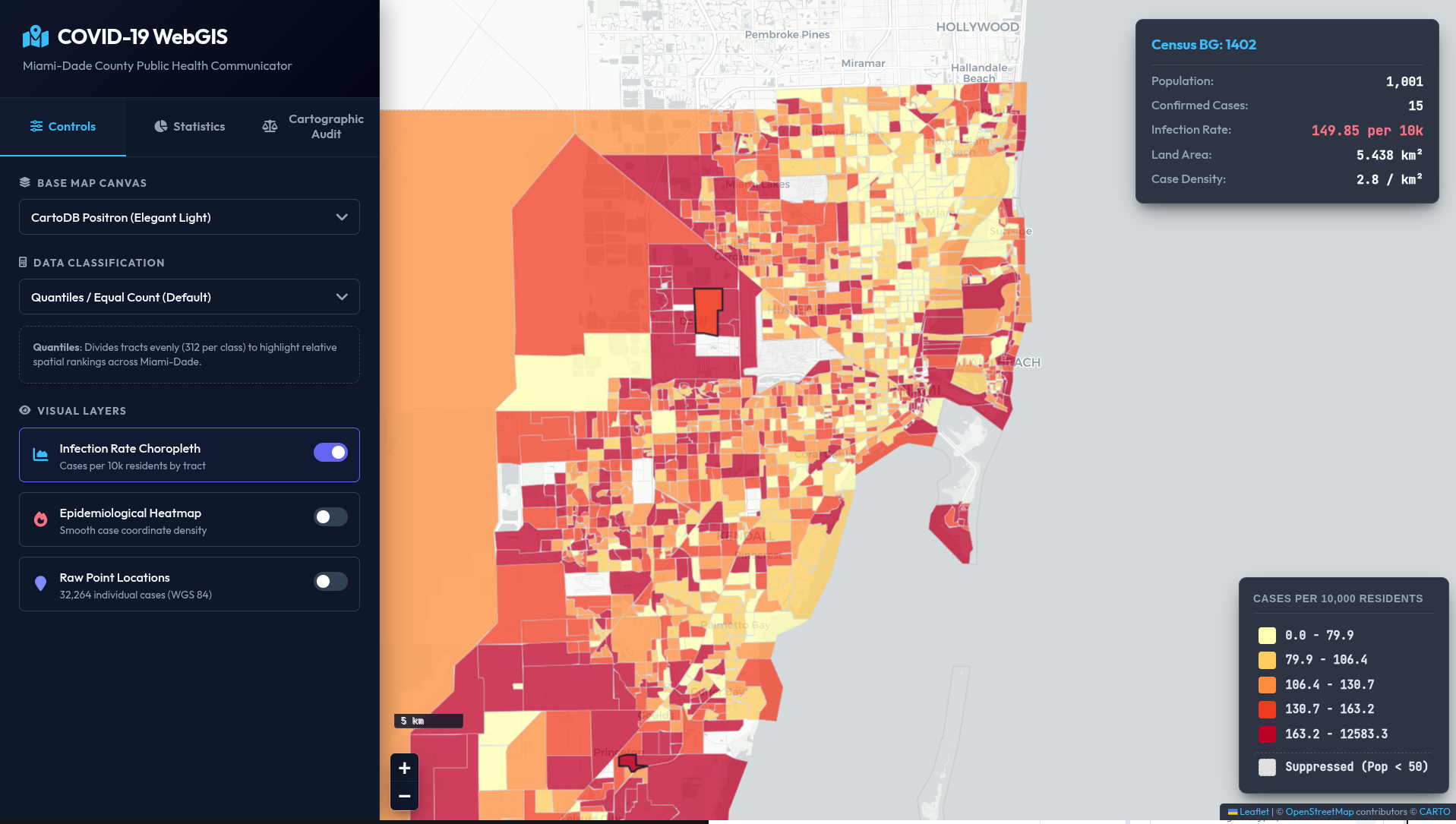}
        \caption{}
        \label{fig:map_agentloom_a}
    \end{subfigure}
    \hfill
    \begin{subfigure}[b]{0.25\textwidth} 
        \centering
        \includegraphics[height=7cm]{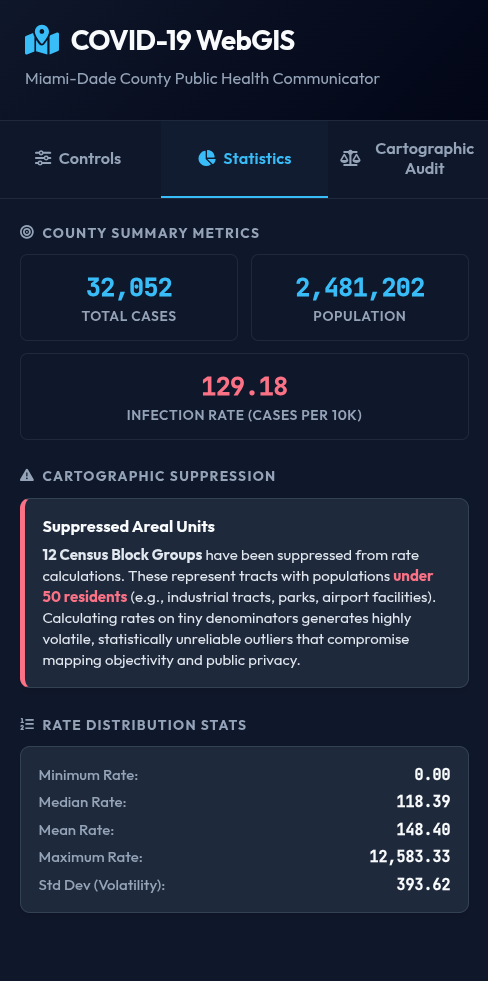}
        \caption{}
        \label{fig:map_agentloom_b}
    \end{subfigure}
    
    \caption{Output from Condition B (Full Dual-Helix). (a) The governed agent successfully enforces domain constraints, calculating normalized rates and rendering a complete legend with source citations. (b) The companion statistics panel surfaces the agent's methodological safeguards: it suppresses census block groups with fewer than 50 residents to prevent volatile rates on small denominators, and reports the county summary metrics together with the resulting rate distribution.}
    \label{fig:map_agentloom}
\end{figure}

The quantitative divergence is most visibly apparent when examining the specific cartographic failures of the unguided model. Left unguided, the LLM frequently generated flawed maps. Figure~\ref{fig:map_baseline} demonstrates an example with flawed visualizations, technical glitches (persistent pop-ups), visual clutter and information overload. This behavior actively mimics the real-world vectors of the COVID-19 infodemic, where decoupled, contextless data visualizations facilitated public misinterpretation. In contrast, the Full AgentLoom framework (Condition B) showed more reliability in adhering to the externalized cartographic rules derived from \citet{mooney_mapping_2020}. By programmatically enforcing the \texttt{webgis:misinformation} and \texttt{webgis:design-guidelines} knowledge nodes prior to code execution, the agent consistently applied population normalization algorithms, utilized sequential palettes, and generated less cluttered UI elements (Figure~\ref{fig:map_agentloom}). A unique characteristics of this particular solution was the suppression of areas with low populations but high concentration of reported infections (e.g. an airport) that would artificially skew the results, as well as prominently documenting this in the interface.

\section{Discussion of the Results} \label{section-discussion}

The previous section demonstrated the feasibility of the proposed framework through two distinct case studies. The first case study described the process of refactoring the FutureShorelines legacy codebase into a modern, modular architecture (Section~\ref{section-futureshorelines}), along with a quantitative experiment demonstrating increased operational reliability of a dual-helix governed agent (Section~\ref{section-futureshorelines-experiment}). The second case study described in Section~\ref{section-case-study-covid} provides further evidence of our approach's generalizability by applying the framework on a cartographic visualization task related to the COVID-19 pandemic and evaluating common cartographic pitfalls. 

\subsection{Addressing the Five LLM Limitations} 

Current LLMs suffer from several limitations ($C1-C5$ in Section~\ref{section-intro-challenge}) that compromise their ability to produce production-level WebGIS software. In this paper, we argued that overcoming these limitations and utilizing LLMs in this context cannot be sufficiently addressed by model capability alone, but requires externalized governance structures. We proposed a dual-helix governance framework that stabilizes LLM-powered agentic execution through two axes: knowledge externalization and behavioral enforcement (Section~\ref{section-dual-helix}). The case studies described in Section~\ref{section-futureshorelines} and Section~\ref{section-case-study-covid} provide multi-dimensional evidence that this governance-based approach successfully mitigates each of the five core LLM limitations we identified in WebGIS contexts. Here, we summarize how each limitation was addressed.

Long-context Limitations ($C1$) can occur when a long codebase, a massive input file or accumulated instructions exceed the attention range of an LLM. We find that this can be mitigated by externalizing structural code patterns and domain facts into a Knowledge Graph (KG). Rather than attempting to process the entire file through attention mechanisms, our approach externalized key structural information to this KG. For example, in our unguided baseline (Condition A), the model frequently struggled to hold the entire 2,432-line legacy codebase in its active attention, and forgot the \texttt{CONFIG} structure established in Step 1 during later steps, reverting to hardcoded values (e.g., writing \texttt{zoom: 12} instead of referencing \texttt{CONFIG.map.zoom}).

We termed the limitation of standard LLM architectures not being able to retain previous decisions or contexts between sessions spanning weeks or months Cross-session forgetting ($C2$). In the FutureShorelines case study, the KG served as a persistent substrate that maintained architectural decisions and domain terminology across four sessions over three days. Unlike traditional sessions in LLM and chatbot interactions that require repeated re-explanation, our governed agent resumed work seamlessly by referencing prior nodes, such as an approved refactoring plan.

LLMs are also not deterministic. This output stochasticity ($C3$) means that the same task may produce different results across different runs. This can manifest in different module structures or architectural patterns across in a code refactoring task, or distinct cartographic choices in web map design across different runs. In the FutureShorelines case study, we note that without governance, event-driven integrations proved fragile; the ungoverned model randomly omitted required payload fields (e.g., missing the \texttt{feet} attribute in a \texttt{scenario-changed} custom event), causing downstream chart components to silently fail. The skills track of the operational implementation of our dual-helix framework provided validated workflow templates that stabilized execution, ensuring consistent event payloads and communication protocols across all modules. Similarly, in the COVID-19 web mapping case study, the unguided baseline exhibited significant cartographic stochasticity. The agent used a random color scheme unsuitable to represent risk in one instance (Figure~\ref{fig:map_baseline}), while it used graduated colors in another. By enforcing domain rules as mandatory behavioral constraints, the Dual-Helix framework effectively eliminated this variance and consistently used graduated colors to represent increasing risk. 

The controlled experiments presented in Section~\ref{section-futureshorelines-experiment} and Section~\ref{section-case-study-covid} highlighted that Instruction Following Failure ($C4$) can be systematically avoided through behavioral constraints embedded in the KG. Standard LLMs often "normalize" domain rules or fall back on their generalized training data when prompt instructions become too complex. During the code refactoring task, ungoverned models incorrectly rounded exact scientific SLR thresholds (e.g., 0.54m to 0.5m), altered precise DOM IDs (\texttt{ej-polygons1} to \texttt{ej-polygons}), and fell back to legacy anti-patterns like \texttt{MutationObserver}. Similarly, in the automated cartography task, the unguided baseline repeatedly failed to follow epidemiological design instructions, e.g. by omitting critical data provenance (source attributions and methodological warnings). Behavioral enforcement automatically caught these violations. By anchoring rules in nodes such as \texttt{behavior:audit-trail} or \texttt{webgis:design-guidelines}, the governed agent enforced domain-specific standards because they were physically integrated into the execution protocol rather than provided as optional suggestions.

Finally, agent behavior is traditionally improved through fine-tuning, which presents Adaptation rigidity ($C5$). Fine-tuning requires massive datasets, takes a long time, and results in opaque parameter changes that cannot be easily audited. The Dual-Helix framework bypasses this rigidity by allowing the system to adapt to new domains structurally. For example, during the code refactoring task,the system adapted to highly specific, project-level accessibility features (e.g., specific screen-reader \texttt{sr-only} \texttt{divs}) by simply encoding them as persistent rules in the Behavior track. In the FutureShorelines case study, our results demonstrated substantial KG growth (from 28 seed nodes to 126 nodes). This learning is observable, auditable (administrators can review and approve changes), and reversible (version control enables rollback) without the need for expensive fine-tuning cycles. This adaptation was also pronounced in the COVID-19 web mapping case study. The base prompt for the unguided agent (Condition A) simply instructed the model to "use your best judgment" to visualize the pandemic data (Supplementary Material Listing~\ref{lst:task-prompt}), which resulted in cartographic errors. Rather than attempting to fine-tune the model on epidemiological mapping standards, the framework adapted instantly. During the deliberate "Build Phase," the Agent Builder simply extracted the cartographic guidelines from \citet{mooney_mapping_2020} and persisted them as JSON graph nodes. This demonstrates that the governance architecture allows an agent to rapidly and auditably adapt to entirely new professional domains without requiring any updates to the underlying foundation model. This longitudinal role also explains the ablation ranking in Section~\ref{section-case-study-covid}: the Knowledge track had the smallest marginal effect in the single-shot cartography task because its primary benefit is cross-session persistence and auditable self-learning, which is exercised by the multi-session refactoring task (the knowledge graph grew from 28 to 126 nodes) rather than within a single generation pass.

\subsection{Structural vs. Informational Strategies} \label{section-discussion-structural}

Our results highlight a fundamental distinction between informational strategies (e.g., prompt engineering, RAG) and structural governance such as the proposed dual-helix framework. While information strategies can describe a rule, they cannot systematically enforce it. Governance artifacts on the other hand form an external computational structure that constrains behavior, rather than merely providing information that the agent may or may not use. Condition B in the FutureShoreline experiment (Static Context in Section~\ref{section-futureshorelines-experiment}) served as a proxy for a "perfect" informational payload, simulating a scenario where all necessary domain rules and project facts were successfully retrieved and supplied to the agent. However, our results demonstrate that simply possessing the information does not guarantee compliance. In a zero-shot or purely retrieval-based setting, mandatory constraints like coordinate system integrity are ultimately treated as advisory tokens appended to a prompt. As the context window grows, the attention weights allocated to these rules can degrade, leading to the instruction-following failure ($C4$). In contrast, structural governance through the Dual-Helix framework externalizes these concerns. By utilizing a 3-track architecture, the system separates the 'What' (domain facts) from the 'How' (executable constraints). While this still operates via prompt injection at the foundation model level, it shifts the paradigm from a single, static context window to dynamic, step-specific structural guidance. Critically, these governance artifacts are persistent, schema-validated, and version-controlled graph nodes that are programmatically assembled into each prompt and not manually authored, ephemeral text prone to degradation as context accumulates. This ensures that compliance is not left to the probabilistic outcome of a well-phrased prompt, but is actively monitored and structurally guided at each step of the agent's operational cycle. For example, rather than hoping a model remembers to apply WCAG accessibility standards after several hours of coding, the governed agent is structurally guided to validate its plan against the persistent Behavior track at key workflow checkpoints.

\subsection{Implications for GIS and Beyond}

Our results demonstrated through two distinct case studies suggest that operational reliability in agentic AI applied on a WebGIS tasks is as much a governance problem as it is a model capacity problem. This reframing has practical implications. The most immediate and straightforward implication for GIS practitioners or small, resource-constrained research teams is that our results suggest that investing in externalized structures may be more effective than waiting for more capable models. This is particularly relevant given the documented GIS curriculum gap, where practitioners often lack formal software engineering and computer science training. By encoding these principles into Behavioral Enforcement protocols, the dual-helix governance framework provides the necessary cyber literacy (see \citep{shook_cyber_2019}) scaffolding to ensure professional-grade output in geospatial tasks requiring computational skills.

While GIScience has researched geo-ontologies for decades to facilitate semantic interoperability \citep{Agarwal01052005}, these were often top-down, static hierarchies. Our results show that knowledge graphs can provide persistent, auditable institutional memory that LLMs inherently lack. This aligns with \citet{couclelis_ontologies_2010}'s argument that an information-based ontology should explicitly build user purpose and object function into geographic constructs. Our framework operationalizes this by shifting these purposes from "advisory" prompt text to structurally governed nodes.

Beyond code refactoring and creating high-quality web maps, these governance principles directly address the emerging paradigm of autonomous GIS, which seeks systems capable of self-generation, self-execution and self-verification \citep{li_giscience_2025}. Rather than relying exclusively on informational retrieval, this framework utilizes the Knowledge Axis to store "validated geoprocessing workflows," directly addressing key gaps regarding data operation and decision-making identified by \citep{li_giscience_2025}. The reliability ceiling explored in this study is a universal challenge for LLM deployment in any high-stakes environment. Notably, \citet{wei_agentic_2026} identify the development of governance frameworks for agentic reasoning as a central open problem, observing that failures in autonomous systems ``may arise from interactions across time and components, making attribution and auditing difficult.'' Our dual-helix framework directly responds to this open problem by providing a concrete, empirically validated governance architecture that addresses exactly these concerns through persistent, auditable, and version-controlled knowledge structures. We presented two case studies that were implemented in different agentic IDEs (Cursor and Google Antigravity) and used different underlying LLMs (\texttt{gpt-5.2} and \texttt{gemini-3.5-flash}). This suggests that the Dual-Helix approach is flexible and generalizable, and could be applied to other domains (e.g., legal technology or bioinformatics) where the gap between "advisory" information and "mandatory" compliance currently prevents production-level implementation. Within the geospatial domain, this governance architecture enables several future use-cases:

\begin{itemize}
    \item \textbf{Automated spatial analysis:} Ensuring consistent methodology by anchoring geoprocessing workflows in validated skill nodes.
    \item \textbf{Systematic documentation:} Generating technical artifacts that strictly follow organizational standards rather than mere linguistic patterns.
    \item \textbf{Data quality assurance:} Mandating that data operations comply with known schemas and non-negotiable topological standards.
    \item \textbf{Multimodal fusion:} Extending governance to tasks highlighted by \citet{ameen_review_2026}, such as automated data annotation, pixel-level grounding in high-resolution imagery, and the integration of unstructured sensor streams for real-time traffic or water pollution monitoring.
    \item \textbf{Professional training:} Creating assistants that consistently apply best practices to bridge curriculum gaps in specialized domains.
\end{itemize}

\subsection{Limitations of this Study} \label{section-discussion-limitation}

While the dual-helix approach demonstrates improvements in operational reliability for WebGIS development, several limitations and potential threats to validity must be considered. These are summarized and listed below:

\begin{itemize}
    \item \textbf{Scope and Generalizability:} This study is fundamentally an architectural framework paper supported by two case studies (code refactoring and automatic web cartography). Broader validation across other diverse GIS tasks (e.g., complex spatial joins and analysis) is necessary to confirm the framework's universal generalizability. 
    \item \textbf{Baseline Subjectivity:} Condition B (Static Context) in the controlled experiment (Section~\ref{section-futureshorelines-experiment}) relied on a manually crafted static prompt. While one might argue this introduces subjectivity, this limitation highlights the exact problem the framework solves: the inherent subjectivity in human-engineered prompts across long-horizon projects is what necessitates the programmatic, objective extraction of context from a Knowledge Graph. Condition B represents a realistic ``best effort'' for informational prompting.
    \item \textbf{Absence of a Retrieval-Augmented Baseline:} Our controlled experiment compared structural governance against an unguided agent and a static-context prompt, but did not include a dynamic Retrieval-Augmented Generation (RAG) condition that retrieves relevant documents at each step. We therefore frame structural governance and retrieval as complementary rather than competing strategies, and leave a direct comparison between the two to future work.
    \item \textbf{Evaluation Mechanisms and LLM Bias:} While our finding regarding Rule Compliance and Domain Accuracy in the first case study were derived from deterministic checks (i.e. verifying the presence of specific API calls or layer IDs), we utilized an LLM-as-a-judge framework for qualitative metrics (like cross-step coherence) that may introduce bias in the evaluation.
    \item \textbf{Token Volume Dynamics:} The dynamic context condition inherently retrieved more tokens than the static prompt. Rather than a confounding bug, we consider this automated, high-density retrieval a feature of structural governance. Automated retrieval bypasses human cognitive limits and provides the right context when needed. Nevertheless, this potential limitation is acknowledged.
    \item \textbf{Statistical Power and Sample Size Constraints:} Due to the substantial API costs and computational latency associated with executing multi-step, autonomous agentic workflows, the experimental trials were constrained ($n=5$ for the refactoring experiment, and $n=3$ per condition for the COVID-19 ablation study). As a result, standard variance-based statistical tests (such as the Kruskal-Wallis H-test, $p=0.15$) did not meet the strict $\alpha=0.05$ threshold for significance. However, in applied software engineering, practical impact often supersedes traditional $p$-value thresholds. The effect sizes observed (a small-to-medium Cohen's $d = 0.46$ for the refactoring task and a large $\epsilon^2 = 0.48$ for the cartography task), combined with qualitative consensus (100.0\% adjacent evaluator agreement for the fully governed cartography), demonstrate a practical difference. The framework successfully and predictably prevents catastrophic architectural and cartographic failures, which is the primary prerequisite for production deployment.
    \item \textbf{The Autonomy vs. Human-in-the-Loop Balance:} Our implementation utilizes a ``Plan-first'' rule requiring human approval. While this might appear contrary to the goal of ``Autonomous GIS,'' role separation and human oversight serve as necessary safeguards against hallucinations (which are often catastrophic in software development), representing a responsible pathway toward fully autonomous GIS. Nevertheless, one must be aware of this distinction when interpreting our study in the context of Autonomous GIS.
    \item \textbf{Upfront Investment:} Constructing the initial governance structure, including the Knowledge Graph, requires a meaningful upfront investment in time and expertise. For small, single-script tasks, this overhead may not be justified compared to standard prompting.
\end{itemize}
\section{Summary and Future Work} \label{section-summary-futurework}

WebGIS development presents unique challenges for AI-assisted workflows, including long development cycles and the need to adhere to strict institutional standards. We identified five core LLM limitations, i.e. long-context constraints, cross-session forgetting, output stochasticity, instruction following failure, and adaptation rigidity, that frequently undermine reliability in these tasks. This paper proposed a dual-helix governance framework that addresses these challenges through structured knowledge externalization and enforceable behavioral constraints.

Our central contribution is the demonstration that structure, not just capability, determines LLM reliability in specialized domains. By externalizing project memory into a persistent Knowledge Graph and mandating compliance through an executable Behavior axis of the helix, this approach allows agents to maintain architectural consistency and scientific integrity. We also presented the operationalization of this framework as a 3-track architecture consisting of Knowledge, Behaviors, and Skills. It is also available as the open-source software AgentLoom (\url{https://doi.org/10.5281/zenodo.17561541}, \citep{guan_agentic-ai_2025}). Through two distinct case studies, we demonstrated the versatility and robustness of this approach. First, we showed that a governed agent can autonomously refactor a monolithic 2,265-line codebase into a modern modular system, reducing code complexity and improving maintainability. Second, in an automated COVID-19 web mapping task, we demonstrated that the framework effectively mitigates critical "infodemic" vulnerabilities, such as failing to normalize data or omitting data provenance, by structurally enforcing established epidemiological cartography rules. A controlled ablation experiment comparing this framework against traditional prompting baselines confirmed that dynamic governance increases operational reliability by drastically reducing trial-to-trial variance. Ultimately, this provides a practical methodological pathway for the GIS community to integrate agentic AI into production workflows without waiting for the next generation of "smarter" models.

While the dual-helix approach provides a foundation for more reliable agentic workflows, several research directions remain to be explored. Most importantly, future work should explore the generalizability of this approach to different domains and benchmark the dual-helix approach against other task-specific GeoAI frameworks (e.g., MapAgent or ShapefileGPT) to explicitly isolate the performance gains of structural governance versus specialized model architectures. While this study utilized a single-agent setting, exploring how the dual-helix structure can coordinate multiple specialized agents (e.g., a geospatial analyst and a UI/UX designer) could further enhance performance for large-scale enterprise GIS projects. Another potential direction is investigating how governance structure can be shared or transferred between related projects, potentially leading to a centralized repository of organizational best practices. Lastly, more research is needed to determine the optimal division of labor between humans and AI in building and refining these governance structures, particularly regarding the validation of autonomously discovered project context nodes.


\section*{Data Availability Statement}
The AgentLoom governance framework is available as open-source software at \url{https://doi.org/10.5281/zenodo.17561541} \citep{guan_agentic-ai_2025}. The automated evaluation pipeline code and experimental results (Jupyter notebooks) used for the controlled experiment described in Section~\ref{section-futureshorelines-experiment} are available at the same repository. The FutureShorelines application source code is not publicly available due to institutional licensing restrictions; however, the governance framework and experimental methodology are fully reproducible with any comparable WebGIS codebase.

\bibliographystyle{apalike}
\bibliography{references}

\clearpage
\section*{Supplementary Material}
\appendix
\setcounter{figure}{0}
\renewcommand{\thefigure}{A\arabic{figure}}

\setcounter{table}{0}
\renewcommand{\thetable}{A\arabic{table}}

\setcounter{listing}{0}
\renewcommand{\thelisting}{A\arabic{listing}}

\section{Case Study: Frontend Refactoring Plan and Summary} \label{supplementary-refactor}

\begin{figure}[H]
\centering
\includegraphics[width=\textwidth]{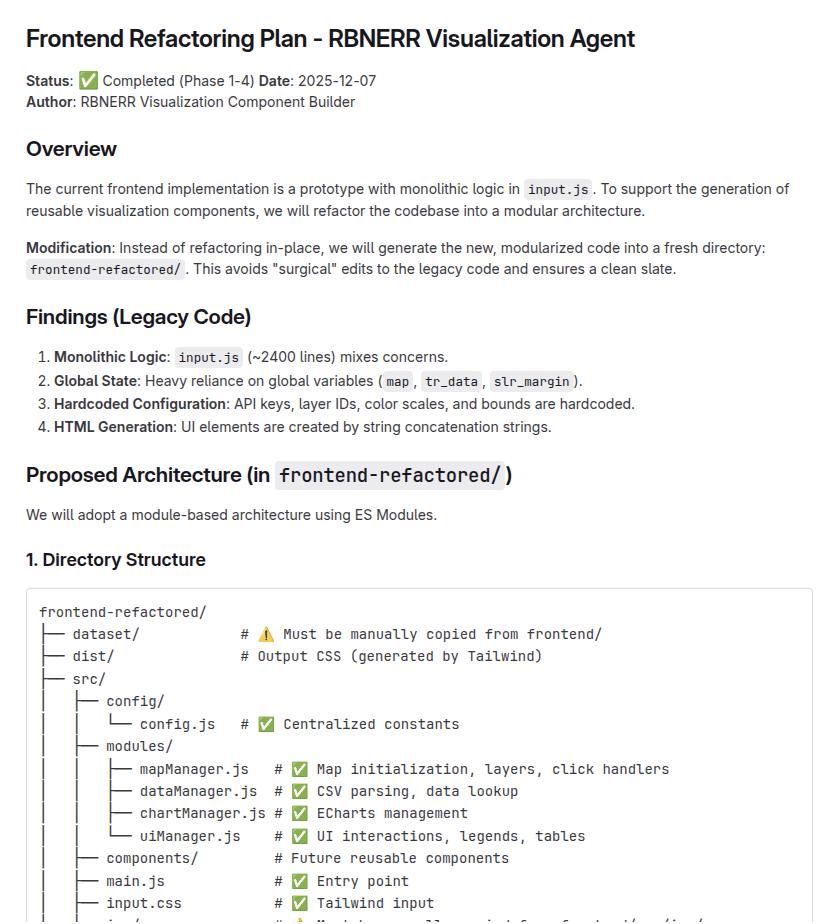}
\caption{Excerpt from the Refactoring Plan generated by the Agent. Note that 1) the agent modified an original plan to override the original codebase and documented this change based on human feedback (see Section~\ref{section-futureshorelines-refactor}) and 2) the agent also documented its completion by changing its status.} 
\label{fig:suppl-refactor}
\end{figure}

\setcounter{figure}{0}
\renewcommand{\thefigure}{B\arabic{figure}}

\setcounter{table}{0}
\renewcommand{\thetable}{B\arabic{table}}

\setcounter{listing}{0}
\renewcommand{\thelisting}{B\arabic{listing}}

\section{Experimental Task Instructions} \label{supplementary-workflow}
All experimental conditions (A, B, and C) received identical task instructions. Steps 1--4 included the instruction followed by the full 2,432-line legacy source file.

\begin{listing}[htbp]
\caption{Step 1: Extract Centralized Configuration Module}
\begin{lstlisting}
Refactor the legacy code into a centralized configuration module.
The legacy application scatters constants, magic numbers, and
configuration values throughout the file as global variables.

Extract these into a single config.js module that exports a CONFIG object.
The module must include:
- Mapbox access token and map settings (style, center, zoom, bounds)
- SLR scenario lookup table (all year x scenario combinations with feet values)
- Color definitions (parcel colors, EJ color stops)
- Layer identifiers (all SLR layers, ocean extent layers, source names)
- Distance tick arrays for charts
- Map bounds coordinates

Replace the legacy 30+ branch if/else chain with a structured data lookup.
Output a complete, self-contained config.js ES6 module.
\end{lstlisting}
\end{listing}

\begin{listing}
\caption{Step 2: Refactor Map Initialization into MapManager Class}
\begin{lstlisting}
Refactor the legacy map initialization, layer setup, and map interaction
code into a MapManager class module. The legacy code has all map logic
inline within a single file with deeply nested callbacks.

The MapManager must:
- Import configuration from the config module created in Step 1
- Initialize Mapbox GL JS map with controls (navigation, scale, attribution, minimap)
- Add all data sources (irl vector tiles, ej polygons, DEM raster)
- Add all layers (parcel outline, parcel highlighted, transect highlighted,
  EJ polygons, EJ highlighted, DEM, point highlighted)
- Handle map click events (parcel selection, transect selection, EJ selection)
- Manage SLR scenario layer switching
- Manage ocean extent layer visibility
- Use event-driven communication (dispatch CustomEvents instead of direct function calls)

Output a complete MapManager ES6 module class.
\end{lstlisting}
\end{listing}

\begin{listing}
\caption{Step 3: Refactor Chart Logic into ChartManager Class}
\begin{lstlisting}
Refactor the legacy ECharts initialization, configuration, and data
transformation code into a ChartManager class module.

The ChartManager must:
- Import configuration from the config module (Step 1)
- Follow the same class-based pattern as MapManager (Step 2)
- Initialize both elevation and slope ECharts instances
- Provide methods to build elevation series data (with MHHW line)
- Provide methods to build slope series data
- Update chart options when parcel/transect selection changes
- Use the same distance_list from config (not hardcoded)
- Include ARIA descriptions for screen readers

Output a complete ChartManager ES6 module class.
\end{lstlisting}
\end{listing}

\begin{listing}
\caption{Step 4: Refactor UI Interactions into UIManager Class}
\begin{lstlisting}
Refactor the legacy UI interaction code into a UIManager class module.
The legacy code has UI logic scattered across 800+ lines with direct DOM
manipulation and no accessibility support.

The UIManager must:
- Import configuration from the config module (Step 1)
- Follow the same class-based pattern as MapManager and ChartManager (Steps 2-3)
- Manage sidebar resize logic
- Handle layer toggle checkboxes and legend updates
- Handle year and sea level sliders (using the SLR lookup from config, NOT if/else)
- Implement search functionality (parcel ID, coordinates, address)
- Handle download buttons (KML, CSV)
- Update parcel and EJ data tables
- Add ARIA labels to all interactive elements
- Add keyboard navigation support (Tab/Shift-Tab, Enter/Space)
- Use event-driven communication (dispatch CustomEvents for cross-module actions)

Output a complete UIManager ES6 module class with full accessibility support.
\end{lstlisting}
\end{listing}

\begin{listing}
\caption{Step 5: Write Refactoring Documentation}
\begin{lstlisting}

Write complete documentation for the refactored modular architecture.
This documentation must accurately reflect the actual modules, classes,
methods, and patterns produced in Steps 1-4.

The documentation must cover:
- Architecture overview (monolithic to modular transformation)
- Module dependency diagram (which module imports what)
- config.js: All CONFIG keys and their types/defaults
- MapManager, ChartManager, UIManager: All public methods
- Event system: All CustomEvents dispatched and consumed, with payload schemas
- Migration guide: Legacy global functions mapped to new module methods
- Accessibility improvements

The documentation MUST use the exact method names, parameter names, event
names, and CONFIG keys from the actual code in Steps 1-4.

Output a complete Markdown documentation file.
\end{lstlisting}
\end{listing}

\setcounter{figure}{0}
\renewcommand{\thefigure}{C\arabic{figure}}

\setcounter{table}{0}
\renewcommand{\thetable}{C\arabic{table}}

\setcounter{listing}{0}
\renewcommand{\thelisting}{C\arabic{listing}}

\newpage
\clearpage
\section{System Prompt Configurations} \label{supplementary-systemprompt}

\begin{listing}
\caption{Full Static Prompt for Condition B that received a fixed, 4,000-token project brief as the system prompt for every step.}
\label{supplementary-systemprompt-b}
\begin{lstlisting}
You are an expert web developer refactoring a legacy geospatial visualization
application at the Rookery Bay National Estuarine Research Reserve (RBNERR).
The task is to transform a 2432-line monolithic JavaScript file (input.js) into
a modular ES6 architecture.
## Legacy System Problems
- Monolithic structure, global variables, and hardcoded values.
- Brittle MutationObserver coupling.
- No accessibility support.
## Target Architecture
- config.js -> mapManager.js -> chartManager.js -> uiManager.js -> main.js.
## Domain Knowledge
- SLR scenarios: baseline, 2030, 2040, 2050, 2100.
- Source IDs: 'irl', 'ej', 'dem', 'composite'.
- Scenario levels: interm-low, interm, interm-high, high.
## Critical Rules
- SLR values must match legacy exactly -- do not round or approximate.
- All layer IDs and source IDs must match the Mapbox style verbatim.
- Vector tile property names are schema-bound -- do not camelCase or rename them.
- Replace MutationObserver with direct event dispatch from sliders.
- DOM element IDs must match the HTML template exactly.
- WCAG 2.1 AA: ARIA labels on all elements, keyboard navigation support.
\end{lstlisting}
\end{listing}

\begin{listing}
\caption{Representative Dynamic System Prompt Snippets (Step 4) from Condition C: Dynamic Context (Dual-Helix)}
\label{supplementary-systemprompt-c}

\begin{lstlisting}
## Role & Context
You are performing step 4: Refactor UI Interactions into UIManager Class.

## Mandatory Constraints (Critical Priority)
--- id: behavior:rbnerr-viz-builder:dom-id-preservation ---
# Behavior: DOM Element ID Preservation
Rule 1: Never Rename Element IDs -- They Are HTML Contract.
Reference IDs exactly: 'ej-polygons1', 'UCF outfalls', 'sl-opac', etc.

--- id: behavior:rbnerr-viz-builder:cross-module-event-contract ---
# Behavior: Cross-Module Event Contract
Rule: CustomEvent payloads are strict contracts.
'scenario-changed' detail must include: { year, level, feet }.

## Required Constraints (High Priority)
--- id: behavior:rbnerr-viz-builder:mutation-observer-replacement ---
# Behavior: MutationObserver Anti-Pattern Replacement
Rule: Do not copy MutationObserver logic. Use direct event dispatch from sliders.

## Accumulated State
- Config keys established in S1: CONFIG.mapbox.token, CONFIG.slr.scenarios.
- Class signatures from S2: MapManager.handleEvent(), MapManager.update().
\end{lstlisting}
\end{listing}

\newpage
\clearpage

\setcounter{figure}{0}
\renewcommand{\thefigure}{D\arabic{figure}}

\setcounter{table}{0}
\renewcommand{\thetable}{D\arabic{table}}

\setcounter{listing}{0}
\renewcommand{\thelisting}{D\arabic{listing}}
\section{Detailed Evaluation Rubric} \label{supplementary-eval}

The scoring framework applies a three-level compliance scale (0--2) across six dimensions.

\begin{table}[h]
\centering
\small
\begin{tabular}{lp{3.5cm}p{8cm}}
\toprule
\textbf{ID} & \textbf{Dimension} & \textbf{Success Criteria (Score = 2)} \\ \midrule
E1 & Domain Accuracy & Exact SLR lookup values (e.g., 0.54, 6.81); exact layer IDs (\texttt{sl-baseline-v3}); exact GIS field names (\texttt{DEMOGIDX\_2}). \\
E2 & Accessibility & ARIA wrappers on canvas elements; keyboard handlers (\texttt{keydown}); implementation of \texttt{tabindex}. \\
E3 & Pattern Consistency & Uses class-based manager pattern; references centralized \texttt{CONFIG}; uses \texttt{CustomEvent} dispatch. \\
E4 & Coherence & Accurate reuse of methods, events, and config keys defined in prior steps. \\
E5 & Rule Compliance & Zero use of \texttt{MutationObserver}; exact preservation of critical DOM IDs (\texttt{ej-polygons1}). \\
E6 & Documentation & Matches actual implementation: correct class names, method signatures, and event names. \\ \bottomrule
\end{tabular}
\caption{Rubric used to evaluate Case study 1}
\end{table}

\subsection{Deterministic Verification Checks}

\begin{itemize}
    \item \textbf{Data Integrity (E1):} Presence of 2030 SLR values (0.54, 0.56, 0.60, 0.61) and 2100 values (6.81, 5.17, 3.63, 2.08).
    \item \textbf{Behavioral Enforcement (E5):} Negative check for the string \texttt{MutationObserver}; positive check for specific element IDs like \texttt{UCF outfalls} and \texttt{ej-polygons1}.
    \item \textbf{Event Contract (E5):} Validation of the \texttt{scenario-changed} event payload containing \texttt{year}, \texttt{level}, and \texttt{feet}.
\end{itemize}

\newpage
\clearpage
\setcounter{figure}{0}
\renewcommand{\thefigure}{E\arabic{figure}}

\setcounter{table}{0}
\renewcommand{\thetable}{E\arabic{table}}

\setcounter{listing}{0}
\renewcommand{\thelisting}{E\arabic{listing}}

\section{COVID-19 Case Study} \label{supplementary-covid}

\begin{listing}
\caption{Exceprt from the "magic activation prompt" from AgentLoom framework's Build phase, instructing the Agent Builder role to parse the existing project directory and create the Webgis Developer agent. }
\label{lst:activation-prompt}
\begin{lstlisting}
# Task: Build Custom AI Agent Using AgentLoom

Read the protocol package:
@agentLoom/phases/INDEX.md

Start with Phase 0 to verify environment setup.

## My Project Definition
**Domain**: web mapping
**Purpose**: to create web maps that communicate covid-19 infections
**Primary users**: project manager

## Domain Role (Custom Role)
**Role name**: WebGIS Developer
**Role ID**: webgis-developer
**Personality**: technical
**Communication style**: technical

**Primary tasks** this role performs:
1. handle geospatial data
2. create web maps
3. design web maps

**Boundaries** (what it won't do):
- Do not edit or read files outside the project folder.

## Knowledge Requirements
**Content sources**:
General web mapping knowledge, and GIS knowledge.

**If you have existing content folders**:
@content-raw/*.geojson - geojson files that will be used as input
@content-raw/mapping_paper.pdf - very important for covid-19 web map failure modes

## Skills & Capabilities

[..omitted for brevity..]
---
BEGIN with Phase 0: Environment Setup
\end{lstlisting}
\end{listing}

\begin{listing}
\caption{Prompt describing the task for the autonomous Execution phase}
\label{lst:task-prompt}
\begin{lstlisting}
Develop a single-file interactive WebGIS application (HTML/JS) using an open source or freely available web mapping library to communicate COVID-19 risk in Miami-Dade County.

## Available Data:

covid_sample_miami_dade.geojson: Point locations of confirmed COVID-19 cases in 2021 January.
block_groups_miami_dade.geojson: Polygon boundaries containing a attributes for population (pop100) and the land area in square meters (aland10) and the COVID-19 case counts for 2021 January.

The files are located in the content-raw folder in the  project folder. The data is authoritative, the source is the Department of Health.

## Requirements:

Use your best judgement and knowledge to decide how to present the COVID-19 data on a web map.

Explore the geojson inputs, and enforce data validation. For example, spatial operations might fail if features have have null, undefined, or invalid geometries.

Process the point and polygon data to create an interactive map visualizing the COVID-19 impact in Miami-Dade County.

Implement an interactive legend that dynamically reflects the data classification.

Add a hover tooltip that displays the relevant block group statistics.

Ensure the map is accessible and ready for public dissemination.

Put the output code in a folder named results in the project directory. Also copy the input geojson files there, which your script should load. . The results will be viewed in a Google Chrome browser.

Make sure that the visualization works.
\end{lstlisting}
\end{listing}

\begin{table}[htbp]
\centering
\caption{Cartographic Reliability Evaluation Rubric, adapted from \citet{mooney_mapping_2020}.}
\renewcommand{\arraystretch}{1.4}
\small
\begin{tabularx}{\textwidth}{@{} >{\hsize=.8\hsize\raggedright\arraybackslash}X >{\hsize=1.06\hsize\raggedright\arraybackslash}X >{\hsize=1.06\hsize\raggedright\arraybackslash}X >{\hsize=1.06\hsize\raggedright\arraybackslash}X @{}}
\toprule
\textbf{Criterion} & \textbf{Pass (2)} & \textbf{Partially OK (1)} & \textbf{Fail (0)} \\
\midrule

\textbf{Aggregation \& Metrics} (D1) & 
Defaults to population-normalized rates. If raw counts or area-based metrics are offered, they are accompanied by explicit statistical disclaimers. & 
Offers normalized rates but defaults to raw counts or non-standard metrics (e.g., cases/area) without providing clear methodological warnings. & 
Restricts users entirely to raw absolute counts for polygon aggregation. \\

\textbf{Classification \& Symbology} (D2) & 
Defaults to robust algorithms (Jenks, quantiles) and perceptually uniform color ramps. User flexibility (e.g., changing class count) is supported and updates correctly. Colors are not biased. & 
Defaults to easily skewed methods (e.g., equal interval) but allows user adjustment, or utilizes sub-optimal but decipherable color palettes or visualizations, but ability to choose correct methods exist. & 
Locks users into statistically invalid classifications, emotionally biased colors, or mathematically incorrect heatmaps with no ability to adjust. \\

\textbf{Layer \& View Management} (D3) & 
Employs mutually exclusive toggles or effective transparency to cleanly separate distinct visualization types (points, heatmaps, choropleths) without occlusion. & 
Allows overlapping layers that create minor visual clutter or partial occlusion, but maintains basic analytical usability. & 
Forces incompatible layers simultaneously (e.g., dense raw points directly over choropleths), causing severe data masking and cognitive overload. \\

\textbf{Spatial Uncertainty} (D4) & 
Explicitly documents spatial limitations (Modifiable Areal Unit Problem, point coordinate fuzziness) within the UI or alongside the metric selection tools. Includes metadata and source. & 
Includes basic dataset metadata but omits discussion of the spatial uncertainty introduced during point-in-polygon aggregation. & 
Completely lacks representation or textual acknowledgment of methodological and spatial limitations. \\

\textbf{Technical Functionality} (D5)& 
UI elements (toggles, classification inputs, metric selectors) execute without errors and update the map dynamically, precisely, and accurately. & 
Contains minor non-breaking errors, slow rendering, or slight graphical artifacts when switching between views or classification methods. & 
Interacting with options causes breaking errors, unresponsive elements, or severe visual rendering failures that impede analysis. \\

\bottomrule
\end{tabularx}
\label{tab:rubric}
\end{table}

\newpage
\clearpage
\setcounter{figure}{0}
\renewcommand{\thefigure}{F\arabic{figure}}

\setcounter{table}{0}
\renewcommand{\thetable}{F\arabic{table}}

\setcounter{listing}{0}
\renewcommand{\thelisting}{F\arabic{listing}}

\section{AgentLoom Substrate Definitions} \label{supplementary-kg-definitions}

\begin{figure}[H]
\centering
\includegraphics[width=\textwidth]{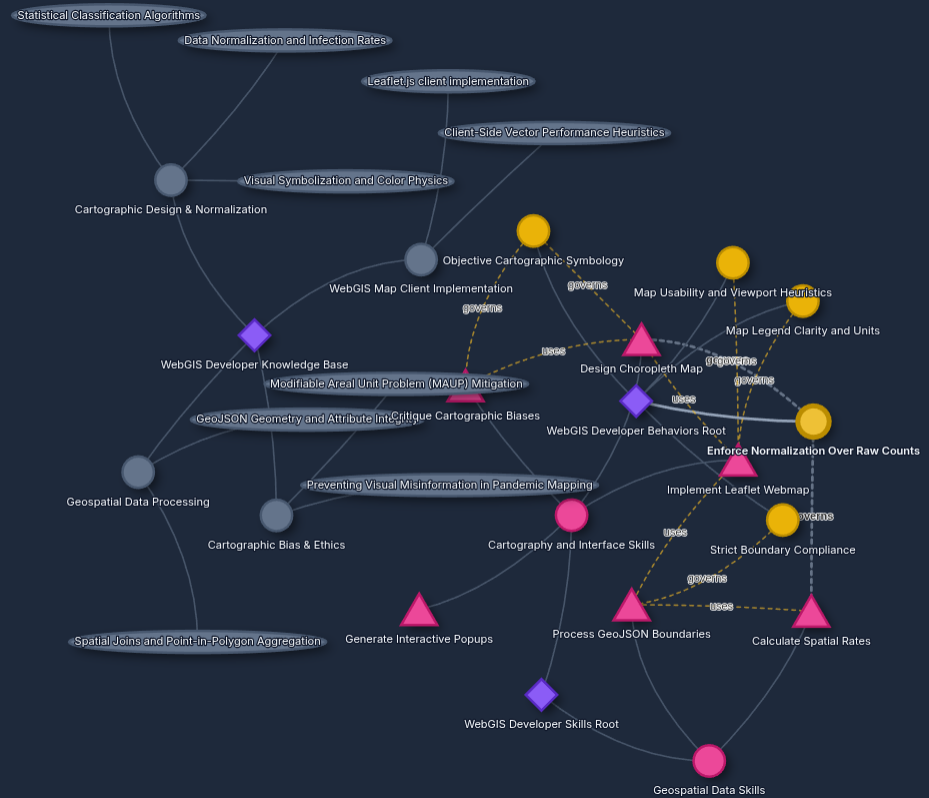}
\caption{Interactive D3.js visualization of the Knowledge Graph substrate generated for the COVID-19 web mapping experiment. The root nodes of the three operational tracks (Knowledge, Behavior, Skill) are shown in purple colors. Behaviors are yellow, while knowledge nodes are grey.} 
\label{fig:suppl-kg-viz}
\end{figure}

The AgentLoom architecture relies on a structured Knowledge Graph (KG) to govern agent actions. The substrate is divided into Knowledge nodes (factual domain information), Behavior nodes (mandatory constraints), and Skill nodes (executable capabilities).  Figure~\ref{fig:suppl-kg-viz} provides a visualization of the interactive D3.js topology generated by the Agent Builder during the COVID-19 cartographic experiment.  The listings in this section provide illustrative examples of how the individual nodes are programmatically defined as JSON objects. The LLM uses the semantic \texttt{description} to retrieve the node contextually, and follows the \texttt{path} to read the expanded instructions within the local Markdown file.

\begin{listing}
\caption{A Knowledge node designed to embed cartographic theory (Mooney \& Juhász, 2020) directly into the agent's context window.}
\label{lst:node-knowledge}
\begin{lstlisting}
{
  "id": "webgis:bias-ethics:infodemic",
  "type": "document",
  "title": "Preventing Visual Misinformation in Pandemic Mapping",
  "parent": "webgis:bias-ethics",
  "description": "Applying Mooney & Juhasz (2020) principles to counter alarming visualizations.",
  "path": "docs/webgis-developer/bias-ethics/visual-misinformation.md"
}
\end{lstlisting}
\end{listing}

\begin{listing}
\caption{Example Behavior node acting as a strict governance protocol. It explicitly links to and governs specific skills, preventing the agent from utilizing raw absolute counts for choropleth rendering.}
\label{lst:node-behavior}
\begin{lstlisting}

{
  "id": "behavior:webgis-developer:prevent-raw-counts",
  "type": "protocol",
  "name": "Enforce Normalization Over Raw Counts",
  "parent": "behavior:webgis-developer:root",
  "description": "Ensuring raw pandemic absolute counts are normalized by population before coloring choropleths.",
  "priority": "high",
  "path": "behaviors/webgis-developer/prevent-raw-counts.md",
  "links": {
    "governs": [
      "skill-calculate-rates",
      "skill-design-choropleth"
    ]
  }
}
\end{lstlisting}
\end{listing}

\begin{listing}
\caption{Example Skill node defining an executable capability. It operates under the constraints established by the Behavior nodes. }
\label{lst:node-skill}
\begin{lstlisting}
{
  "id": "skill-calculate-rates",
  "type": "skill",
  "name": "Calculate Spatial Rates",
  "category": "webgis-developer",
  "parent": "skill:webgis-developer:data-skills",
  "description": "Point-in-polygon aggregation and normalized rate computation per 10,000 residents.",
  "path": "agents/skills/webgis-developer/skill-calculate-rates.md",
  "created": "2026-05-22",
  "links": {
    "uses": [
      "skill-process-geojson"
    ]
  }
}
\end{lstlisting}
\end{listing}

\newpage
\clearpage
\subsection{Markdown Payload Examples}

Each JSON node in the substrate links to a physical Markdown file that contains the expanded instructions, constraints, and algorithmic definitions parsed by the LLM during execution. The following listings display the actual contents of the Markdown files referenced by the JSON nodes above.

\begin{listing}
\caption{Knowledge Node Payload: visual-misinformation.md}
\label{lst:md-knowledge}
\begin{lstlisting}
---
type: document
category: webgis-developer
id: webgis:bias-ethics:infodemic
parent: webgis:bias-ethics
---
# Preventing Visual Misinformation in Pandemic Mapping
## Overview
Web maps are highly powerful visualization tools that are naively [...omitted for brevity...]
## Key Concepts
- **Infodemic**: The rapid spread of accurate and inaccurate [...omitted for brevity...]
- **Cartographic Responsibility**: The ethical obligation of [...omitted for brevity...]
- **DRIP (Data Rich but Information Poor)**: A common [...omitted for brevity...]
- **Visual Misinformation**: The unintended creation of false impressions or inaccurate geographical knowledge due to poor map design, incorrect symbology, or lack of normalization.
## Details
Mooney & Juhasz (2020) critique the widespread misuse of web maps [...omitted for brevity...]
To counter this visual "infodemic", the WebGIS Developer Agent operates under these guidelines:
1. **Never Map Absolute Counts**: Always normalize case metrics by population. Choropleth maps of absolute cases are simply population density maps.
2. **Standardize Aggregation Scales**: Avoid mixing different spatial scales on the same map layer (e.g. plotting Italy as a single dot while showing the US segmented by counties, as seen on early HealthMaps).
3. **Represent Lack of Data/Uncertainty**: Do not paint areas with missing data or laggy testing with a flat "zero cases" indicator. Zero cases should represent verified absence of cases; un-surveyed regions must be visually represented with a distinct, labeled "No Data Available" pattern.
4. **Use Accessible and Objective Symbology**: Color gradients must represent statistical data neutrally. Saturated, scary emotional scales must be replaced with sequential, colorblind-safe sequential ramps.
5. **Ensure Usability**: Standard mapping components (clear legends with exact brackets, scale bars, clear source attributions) are mandatory to avoid cognitive overload.
\end{lstlisting}
\end{listing}

\begin{listing}
\caption{Behavior Node Payload: prevent-raw-counts.md}
\label{lst:md-behavior}
\begin{lstlisting}

---
type: protocol
category: webgis-developer
id: behavior:webgis-developer:prevent-raw-counts
priority: high
---
# Enforce Normalization Over Raw Counts
## Purpose
Ensures absolute pandemic case counts are normalized by population denominators before visual coloring is applied, preventing misleading "population density maps" and cartographic distortions.
## Priority: High
In epidemiological visualizations, failing to normalize counts is a fatal cartographic error that leads to false hotspots, public panic, and visual misinformation.
## Rules
### Rule 1: Mandate Denominators
**What**: The agent must reject mapping any raw absolute disease count unless it is divided by a resident population denominator (`pop100`).
**Why**: Absolute counts are proportional to local populations; normalization isolates the true geographic risk of infection.
**Applies to**: `skill-calculate-rates`, `skill-design-choropleth`
### Rule 2: Constant Base Selection
**What**: Enforce the calculation of infection rates using a standardized base of 10,000 residents.
**Why**: Small fractions (e.g. 0.0079) are cognitively difficult for casual map users; scaling by 10,000 yields legible integers (e.g. 79.4 cases per 10,000).
**Applies to**: `skill-calculate-rates`
### Rule 3: Exclude Volatile Outliers
**What**: Suppress rate calculations in tracts with resident populations below 50.
**Why**: Small denominators yield highly volatile, statistically unreliable rates that skew classification breaks.
**Applies to**: `skill-calculate-rates`
## Governs
**Skills**:
- `skill-calculate-rates`: Dictates the math formula and filters.
- `skill-design-choropleth`: Controls the input properties mapped to colors.
\end{lstlisting}
\end{listing}

\begin{listing}
\caption{Skill Node Payload: skill-calculate-rates.md}
\label{lst:md-skill}
\begin{lstlisting}
---
type: skill
category: webgis-developer
id: skill-calculate-rates
parent: skill:webgis-developer:data-skills
implementation: rule-based
---
# Calculate Spatial Rates
## Purpose
Aggregates individual case points into boundary polygons and computes normalized rates (cases per 10,000 residents) using population denominators.
## When to Use
Triggered when raw coordinate cases and census boundary files must be spatial-joined and normalized.
## Methodology
This skill uses deterministic logic implemented in `scripts/calculate_rates.py`.
### Algorithm
- Reads boundaries GeoJSON and case events GeoJSON.
- Executes Point-in-Polygon spatial joins to count cases per boundary polygon.
- Computes `(cases / pop100) * 10000` for each polygon.
- Excludes boundaries with population below 50 to avoid high outlier variance.
## Example
**Input**:
- Boundaries: `content-raw/block_groups_miami_dade.geojson`
- Cases: `content-raw/covid_sample_miami_dade.geojson`
**Command**: `python3 scripts/calculate_rates.py --boundaries content-raw/block_groups_miami_dade.geojson --cases content-raw/covid_sample_miami_dade.geojson --pop-field pop100 --out output/enriched_boundaries.geojson`
**Output**: Enriched GeoJSON features with:
"properties": {
  "fid": 879,
  "pop100": 1385,
  "cases": 11,
  "rate": 79.42
}
\end{lstlisting}
\end{listing}

\end{document}